\documentclass[10pt]{article} % For LaTeX2e
\usepackage[accepted]{tmlr}
\usepackage{amsmath,amsfonts,bm}

\def\eqref#1{equation~\ref{#1}}
\def\1{\bm{1}}

\DeclareMathAlphabet{\mathsfit}{\encodingdefault}{\sfdefault}{m}{sl}
\SetMathAlphabet{\mathsfit}{bold}{\encodingdefault}{\sfdefault}{bx}{n}

\DeclareMathOperator{\sign}{sign}

\definecolor{cvprblue}{rgb}{0.21,0.49,0.74}
\definecolor{cvprred}{rgb}{0.74,0.21,0.49}
\usepackage[breaklinks,colorlinks,citecolor=cvprblue,linkcolor=cvprred]{hyperref}

\usepackage{url}
\usepackage{graphicx}
\usepackage{tabularx}
\usepackage{booktabs}
\usepackage{multirow}
\usepackage{subcaption}
\usepackage{enumitem}

\usepackage{algorithm}
\usepackage[noend]{algpseudocode}

\makeatletter
\def\BState{\State\hskip-\ALG@thistlm}
\makeatother

\title{Reproducibility Study on Adversarial Attacks Against Robust Transformer Trackers}

\author{\name Fatemeh Nourilenjan Nokabadi \email fatemeh.nourilenjan-nokabadi.1@ulaval.ca \\
      IID, Université Laval $\&$ Mila
      \AND
      \name Jean-François Lalonde \email jflalonde@gel.ulaval.ca \\
      IID, Université Laval
      \AND
      \name Christian Gagné \email christian.gagne@gel.ulaval.ca\\
      \addr IID, Université Laval \\
      Canada CIFAR AI Chair, Mila}

\def\month{05}  % Insert correct month for camera-ready version
\def\year{2024} % Insert correct year for camera-ready version
\def\openreview{\url{https://openreview.net/forum?id=FEEKR0Vl9s}} % Insert correct link to OpenReview for camera-ready version

\begin{document}

\maketitle

\begin{abstract}
New transformer networks have been integrated into object tracking pipelines and have demonstrated strong performance on the latest benchmarks. This paper focuses on understanding how transformer trackers behave under adversarial attacks and how different attacks perform on tracking datasets as their parameters change. We conducted a series of experiments to evaluate the effectiveness of existing adversarial attacks on object trackers with transformer and non-transformer backbones. We experimented on 7 different trackers, including 3 that are transformer-based, and 4 which leverage other architectures. These trackers are tested against 4 recent attack methods to assess their performance and robustness on VOT2022ST, UAV123 and GOT10k datasets. Our empirical study focuses on evaluating adversarial robustness of object trackers based on bounding box versus binary mask predictions, and attack methods at different levels of perturbations. Interestingly, our study found that altering the perturbation level may not significantly affect the overall object tracking results after the attack. Similarly, the sparsity and imperceptibility of the attack perturbations may remain stable against perturbation level shifts. By applying a specific attack on all transformer trackers, we show that new transformer trackers having a stronger cross-attention modeling achieve a greater adversarial robustness on tracking datasets, such as VOT2022ST and GOT10k. Our results also indicate the necessity for new attack methods to effectively tackle the latest types of transformer trackers. The codes necessary to reproduce this study are available at \href{https://github.com/fatemehN/ReproducibilityStudy}{https://github.com/fatemehN/ReproducibilityStudy}.

\end{abstract}

\section{Introduction}

Adversarial perturbations deceive neural networks, leading to inaccurate outputs. Such adversarial attacks have been studied for vision tasks ranging from image classification~\citep{mahmood_robustness_2021, shao_adversarial_2022} to object segmentation~\citep{gu_segpgd_2022} and tracking~\citep{guo_spark_2020, jia_robust_2020, yan_cooling-shrinking_2020, jia_iou_2021}. In this context, transformer-based networks have surpassed other deep learning-based trackers~\citep{li_siamrpn_2019, zhu_distractor-aware_2018}, showing a very robust performance on the state-of-the-art benchmarks~\citep{kristan_tenth_2023}. However, the adversarial robustness of these trackers has not been thoroughly studied in the literature. First, transformer trackers relied on relatively light relation modeling~\citep{chen_transformer_2021}, using a shallow feature extraction and fusion modeling. Based on a mixed attention module, the MixFormer~\citep{cui_mixformer_2022} expanded the road for deeper relation modeling. Consequently, the Robust Object Modeling Tracker (ROMTrack)~\citep{cai_robust_2023} proposed variation tokens to capture and preserve the object deformation across frames. Using transformers, especially those with deep relation modeling~\citep{cui_mixformer_2022, cai_robust_2023}, the object tracker backbones made these models robust to many existing attack approaches~\citep{guo_spark_2020, jia_robust_2020}. Indeed, the underlying concept of adversarial attacks against object trackers is to manipulate the tracker's output. By omitting the multi-head pipelines and substituting them with the transformer backbones~\citep{cui_mixformer_2022, cai_robust_2023}, the tracker's output no longer contains object candidates, classification labels and/or regression labels, which previously were typical targets for attacks. As a result, it is not straightforward to transfer adversarial attacks dealing with classification or regression labels~\citep{guo_spark_2020, jia_robust_2020} on these new transformer trackers. The question also remains on whether other transferable attacks~\citep{jia_iou_2021, yan_cooling-shrinking_2020} can represent a sufficient challenge to transformer trackers. 

%\todo{make sure the number of approaches/attacks/etc. match in the abstract, intro, conclusion.}

This paper presents a study on the reproducibility of existing attack approaches for transformer trackers. We aim to recreate the attack outcomes on transformer trackers using the VOT2022ST~\citep{kristan_tenth_2023}, UAV123~\citep{mueller_benchmark_2016}, DAVIS2016~\citep{perazzi_benchmark_2016} and GOT10k~\citep{huang_got-10k_2021} datasets following two different evaluation protocols, namely \emph{anchor-based short-term tracking} and \emph{One Pass Evaluation (OPE)} protocol. We focus on transformer trackers susceptible to adversarial attacks on their prediction outputs, including the object bounding box and binary mask. Then, we analyzed two white-box attacks on a transformer tracker by varying the perturbation levels and checked its vulnerability to different levels of noise. For the black-box setting, we conducted a similar experiment to assess the attack performance on various noise levels by changing the upper bound of the added noise. In addition, we tested a set of transformer and non-transformer trackers before and after applying the adversarial attacks to discuss the role of transformers in boosting the visual tracking robustness and adversarial robustness. In short, our contributions can be summarized as follows:
\begin{enumerate}[nosep,left=0pt]
    \item We extend the application of adversarial attacks, originally designed for non-transformer trackers like SPARK and RTAA, to assess their effectiveness against transformer-based trackers.
    \item We thoroughly evaluate the adversarial robustness of transformer-based trackers across various output scenarios, perturbation levels, changes in upper bounds, and in comparison to the non-transformer trackers.
\end{enumerate}

\section{Related Work}

\subsection{Visual Object Trackers}

For decades, the tracking task in computer vision has been extensively explored, taking into account various factors and issues. Before the advent of transformers, deep learning-based trackers achieved notable success by leveraging correlation in the form of Siamese networks~\citep{li_siamrpn_2019, zhu_distractor-aware_2018}, and discriminative pipelines for trackers~\citep{bhat_learning_2019, danelljan_probabilistic_2020}. However, new correlation modules in object tracker backbones are being devised through the emergence of transformers. By using transformers in different architectures~\citep{chen_transformer_2021,cui_mixformer_2022, chen_high-performance_2023, cai_robust_2023}, object trackers are capable of inferring object bounding box, binary mask and a prediction score with high robustness values over tracking benchmarks~\citep{kristan_tenth_2023,mueller_benchmark_2016}. These promising results though need to be revisited by assessing transformer trackers in handling the adversarial perturbations. 

The first transformer tracker, TransT~\citep{chen_transformer_2021}, used the cross-attention and self-attention blocks to mix features of the moving target and the search region of the tracker. TransT presents a multi-head pipeline with classification and regression heads, unlike other transformer trackers. In TransT-SEG~\citep{chen_high-performance_2023}, the segmentation head is included in the pipeline. The multi-head pipelines follow the Siamese-based trackers~\citep{li_siamrpn_2019} in dividing each task from target classification to discriminative tracking processing into individual blocks and fusing the results at the end of the tracker structure. Some light relation modeling layers called the Ego Context Augment (ECA) and Cross Feature Augment (CFA) are introduced by TransT to infer the output from combining the multi-head outputs. 
Next, the MixFormer~\citep{cui_mixformer_2022} introduced Mixed Attention Module (MAM) to jointly extract and relate the information from video frames for the object tracking task. By attaching the MixFormer~\citep{cui_mixformer_2022} tracker to the AlphaRefine~\citep{yan_alpha-refine_2021}, MixFormerM~\citep{kristan_tenth_2023} can provide a binary object mask per frame for mask oriented evaluations~\citep{kristan_tenth_2023}. The One-Stream Tracking (OSTrack)~\citep{ye_joint_2022} developed a tracking pipeline that jointly extracts features and models the relation between the search region and the template by bidirectional information flows. In contrast, with the Attention in Attention (AiA) mechanism, the AiATrack~\citep{gao_aiatrack_2022} suggested a three stream framework with long-term and short-term cross-attention modules for relation modeling. Following the further relation modeling, the Robust Object Modeling Tracker (ROMTrack)~\citep{cai_robust_2023} is proposed to enable the interactive template learning using both self-attention and cross-attention modules. The ROMTrack has two main streams to learn discriminative features from hybrid (template and search) and inherent template. The newly introduced variation tokens enable ROMTrack with heavier relation modeling rather TransT~\citep{chen_transformer_2021} and MixFormer~\citep{cui_mixformer_2022}. The variation token carries the contextual appearance change to tackle object deformation in visual tracking task. 

\subsection{Adversarial Attacks Against Trackers}

%Depending on the attack approach, the adversarial loss is different and some elements of the tracker's output are needed to obtain the total loss and produce the final adversarial frame.

The adversarial attack has been proposed in white-box~\citep{guo_spark_2020,jia_robust_2020} or black-box~\citep{jia_iou_2021} attack settings. In black-box attacks, the perturbations are generated without relying on the tracker's gradients, whereas in white-box attacks, adversarial losses are backpropagated through the networks to create the adversarial frame patches (search or template regions). Adversarial attacks against object trackers adopt tracker outputs such as object candidates or classification labels as an attack proxy to generate the adversarial perturbations. For instance, spatial-aware online incremental attack (SPARK)~\citep{guo_spark_2020} creates perturbations by manipulating the classification labels and Intersection of the Union (IoU)~\citep{jia_iou_2021}. It is developed to mislead trackers in providing an altered object bounding box based on the predicted bounding box. In Robust Tracking against Adversarial Attack (RTAA)~\citep{jia_robust_2020}, both classification and regression labels are used to generate the adversarial samples, similar to SPARK~\citep{guo_spark_2020}.  In the RTAA~\citep{jia_robust_2020} algorithm, the positive gradient sign is used to generate the adversarial frames. However, in SPARK, the gradient direction is set to negative following the decoupling of the norm and the direction of gradients in white-box attacks~\citep{rony_decoupling_2019}. Based on the decoupling direction and norm for efficient gradient-based attacks, the direction of the gradient is set in such a way that the generated perturbation has a smaller norm value and greater impact on the results. Some other attacks, such as the Cooling-Shrinking Attack (CSA)~\citep{yan_cooling-shrinking_2020} is developed specifically to impact the output of Siamese-based trackers. In CSA~\citep{yan_cooling-shrinking_2020}, two GANs are trained to cool the hottest regions in the final heatmap of Siamese-based trackers and shrink the object bounding box. Due to dependency on the Siamese-based architecture and loss function, the generalization of the CSA attack~\citep{yan_cooling-shrinking_2020} for other scenarios is harder. The black-box attack, called IoU attack~\citep{jia_iou_2021}, adds two types of noise into the frame to make the tracker predict another bounding box rather than the target bounding box. By considering the object motion in historical frames, the direction of added noise is adjusted according to the IoU scores of the predicted bounding box. 

% %% Rewrite this para

% The first level of the noise is in the tangential direction and for the first four steps, it is added to the frame. Next, in the normal direction, the noise is injected into the frame without any limitations on the steps. The limitations of adding the second noise, is predicting another bounding box with the IoU lower than a threshold or surpassing a higher bound of the noise. 

Before the emergence of vision transformers, some object trackers typically integrated several heads, each assigned to specific vision tasks like classification or segmentation~\citep{li_siamrpn_2019, li_high_2018, zhu_distractor-aware_2018, danelljan_probabilistic_2020}. These deep features were fused to predict the final object position and size, with regression computed over features. In these models, the ultimate decision was made at the network's end based on a set of object candidates. This architectural setup presented various opportunities for crafting adversarial attacks~\citep{guo_spark_2020, jia_robust_2020} against these models, including manipulation of object candidates, object probabilities, and other features, thereby exploiting vulnerabilities in the system. Vision transformers facilitate the integration of features within the deep architecture, enabling direct prediction of the final output, without exposing intermediate outputs that were previously exploitable for attacks. Consequently, white-box attacks utilizing these intermediate outputs (namely, object candidates and their labels) to compute the adversarial loss are no longer applicable to the new transformer backbones via the transformer's gradients themselves. Although they can be transferred in a black-box way (i.e., making adversarial samples with other backbones or other losses), our focus is to employ the transformer gradients in generating adversarial examples in a white-box setting.

\section{Object Trackers and Adversarial Attacks}

In this section, we briefly review the transformer trackers used in our experiments. Also, we explain the adversarial attack methods which are used to attack transformer trackers. The codes and networks of all of the investigated models are publicly available. The implementations of every tracker and every attack approach are the official repository announced by authors. We are also using the fine-tuned and released networks from the authors of the original works. 

\subsection{Object Trackers}

For the object trackers, we considered three types of the robust single object trackers with transformer backbone, two types of the  Siamese-based trackers and two discriminative trackers as follows.

\paragraph{TransT and TransT-SEG} In our studies, we used both Transformer Tracker (TransT)~\citep{chen_transformer_2021} and TransT with mask prediction ability (TransT-SEG)~\citep{chen_high-performance_2023}. By two discriminative streams and a lightweight cross-attention modeling in the end, the TransT introduced the first transformer-based tracker. Similar to the Siamese-based trackers~\citep{li_high_2018,li_siamrpn_2019}, the TransT tracker has two convolutional streams to extract features of the template and the search regions. By proposing Ego-Context Augment (ECA) and Cross-Feature Augment (CFA) modules, a feature fusing network is developed to effectively fuse the extracted features. The ECA infers output as
\begin{equation}
    X_\mathrm{ECA} = X + \text{MultiHead}(X+P_x,\, X+P_x,\,X) \,,
\end{equation}
where ``MultiHead'' represents a multi-head attention module~\citep{vaswani_attention_2017} and $P_x$ is the spatial positional encoding. On the other hand, the CFA module is defined as a Feed-Forward Network (FFN) attached to an ECA module. This FFN includes two linear transformations with a ReLU in between. The CFA's output is computed as:
\begin{align}
\tilde{X}_\mathrm{CF} & = X_q + \text{MultiHead} (X_q+P_q,\,X_{kv}+P_{kv},\,X_{kv}),\nonumber \\
X_\mathrm{CFA} & = \tilde{X}_\mathrm{CF} + \mathrm{FFN}(\tilde{X}_\mathrm{CF}),
\end{align}
where $X_q$ is the input, and $P_q$ is the spatial positional encoding of $X_q$. The input of the cross branch is $X_{kv}$ and its spatial positional encoding is $P_{kv}$. Using two ECAs and two CFAs, the extracted features of the template and search regions are first fused with themselves by ECA and with each other by CFA. Next, another cross attention (CFA) is used to integrate the outputs of the two streams. In the final prediction head, the regression and classification labels are generated to determine the target bounding box by finding the maximum score. The TransT-SEG~\citep{chen_high-performance_2023} has a segmentation branch which uses the template vectors corresponding to the middle position of the template region to obtain an attention map via a multi-head attention module. The object binary mask is, then, predicted by fusing the low-level feature pyramid~\citep{lin_feature_2017} of the search region.

\paragraph{MixFormer and MixFormerM} The MixFormer~\citep{cui_mixformer_2022} is based on a flexible attention operation named Mixed Attention Module (MAM) to interactively exploit features and integrate them in a deep layer of the tracker. The MixFormer coupled with the AlphaRefine network has been proposed for the VOT2022 challenge~\citep{kristan_tenth_2023} as MixFormerM. It enables the original tracker to provide the object mask as an extra output. In our experiments, we tested both MixFormer and MixFormerM trackers. In the MixFormer tracker, the three stages of mixed attention is the core design named MAM which extracts and fuses the features of the template and the search regions. Given the target key, query and value $(k_t, q_t, v_t)$ and the search key, query and value $(k_s, q_s, v_s)$, the mixed attention is defined as:
\begin{align}
    k_m & = \text{contcat}(k_t, k_s), \quad v_m = \text{concat}(v_t, v_s), \nonumber \\
\text{Attention}_t & = \text{softmax}(q_t k_m^T / \sqrt{d})\, v_m, \quad
    \text{Attention}_s = \text{softmax}(q_s k_m^T / \sqrt{d})\, v_m \,,
\end{align}
where $d$ is the key dimension. The Attention$_t$ and Attention$_s$ are the attention maps of the target and search regions. The target and search tokens are, then, concatenated and linearly projected to the output as the mixed attention output. To reduce the computational cost of MAM, the unnecessary cross-attention between the target query and search region is pruned by using the asymmetric mixed attention scheme. It is defined as follows:
\begin{align}
    & \text{Attention}_t = \mathrm{softmax} (q_t k_t^T / \sqrt{d})\, v_t, \quad
    \text{Attention}_s = \mathrm{softmax} (q_s k_m^T / \sqrt{d})\, v_m \,.
\end{align}

The $\text{Attention}_t$ is updated to efficiently avoid the distractors in the search region by fixing the template tokens in the tracking process. In MixFormer, the Score Prediction Module (SPM) is developed by two attention blocks and a three-layer perceptron to identify the target confidence score and update a more reliable online template.

\paragraph{ROMTrack} The ROMTrack~\citep{cai_robust_2023} is developed to generalize the idea of MixFormer~\citep{cui_mixformer_2022} by providing the template learning procedure. The template feature is processed both in self-attention (inherent template) and cross-attention (hybrid template) between template and search regions. This mixed feature avoids distraction in challenging frames and provides a more robust performance compared to TransT and MixFormer. The backbone of ROMTrack is a vision transformer~\citep{dosovitskiy_image_2020} as an object encoder and a prediction head which is a fully convolutional center-based localization head~\citep{Zhou_objects_2019}. Two essential elements of ROMTrack are variation tokens and robust object modeling. The variation tokens are developed to handle the object appearance change and deformation in the tracking process. Considering $F_k^t$ as the output features of $k$-th encoder in frame $I_t$ and $ht_{k,t}$ is the hybrid template part of $F_k^t$, a variation token is defined as
\begin{align}
    & vt_{k,t} = ht_{k,t-1}, \label{eq:5} \\
    & F_{k+1}^t = \text{ObjectEncoder}_{k+1} (\text{concat} (vt_{k,t}, F_k^t)) \,,
    \label{eq:6}
\end{align}
where $F$ represents the output features, $k$ is the encoder index and $t$ denotes the $t$-th frame. In Equation~\ref{eq:5}, ROMTrack saves the hybrid template in the variation token and by Equation~\ref{eq:6}, it embeds the hybrid template into the output feature. The other essential element of ROMTrack is the robust object modeling containing four parts: inherit template $it$, hybrid template $ht$, search region $sr$ and variation tokens $vt$. The inherit template is a self-attention module on the linear projected feature $(q_{it}, k_{it}, v_{it})$ to learn the pure template features $A_{it}$. The hybrid template features and search regions features are obtained via cross-attention. Considering the triplet of cross-attention $(q_z, k_z, v_z)$ as follows:
\begin{align}
    & q_z = [q_{ht}, q_{sr}]\,, \nonumber \\
    & k_z = [k_{vt}, k_{it}, k_{ht}, k_{sr}]\,, \nonumber \\
    & v_z = [v_{vt}, v_{it}, k_{ht}, v_{sr}] \,,
\end{align}
where the different parts of features $(it, ht, sr, vt)$ are rearranged and concatenated to create the cross-attention input. The output of the cross-attention is obtained as:
\begin{equation}
    A_z = \mathrm{softmax}(q_z k_z^T / \sqrt{d})\, v_z \,.
\end{equation}

The features of inherit template and variation tokens are fused effectively with the hybrid template and search region tokens during the tracking process to keep the tracker updated about object deformation and appearance change.

\paragraph{SiamRPN} By integrating the Siamese network for tracking task and Region Proposal Network (RPN) for detection, the SiamRPN tracker~\citep{li_high_2018} is developed to predict a list of object candidates per frame. The Siamese network contains two branches for feature extraction of the template $\varphi(z)$ and search $\varphi(x)$ regions. Then, the region proposal network with a pairwise correlation and supervision, split the feature vectors into the classification $[\varphi(z)]_\mathrm{cls}, [\varphi(x)]_\mathrm{cls}$ and regression $[\varphi(z)]_\mathrm{reg}, [\varphi(x)]_\mathrm{reg}$ vectors. The template feature maps $[\varphi(z)]_\mathrm{cls}, [\varphi(z)]_\mathrm{reg}$ serve as the kernel for computing the correlation as follows:
\begin{align}
    & A^{cls} = [\varphi(z)]_\mathrm{cls} \star [\varphi(x)]_\mathrm{cls} \,, \nonumber \\
    & A^{reg} = [\varphi(z)]_\mathrm{cls} \star [\varphi(z)]_\mathrm{reg} \,,
\end{align}
where $\star$ is the convolution operator. Using a softmax operation, the highest score for classification vector is chosen as the final target class and its corresponding coordinates in the regression vector determine the target bounding box.

\textbf{DaSiamRPN}\quad In DaSiamRPN tracker~\citep{zhu_distractor-aware_2018}, a distractor-aware strategy is taken to improve the SiamRPN tracker~\citep{li_high_2018} performance. In this strategy, the negative samples of object candidates(i.e., proposals) are considered distractor objects. Using Non Maximum Suppression (NMS), a set of distractors is selected to determine the target proposal as the candidate with the highest score. The rest of the set is considered as the distractors and re-ranked by a distractor-aware objective function to reduce their influence in the tracking process.

%%The Discriminative Model Predication(DiMP) tracker~\citep{bhat_learning_2019} adopts a discriminative pipeline that exploits the background information while benefiting from the means of updating the target model. Unlike, Siamese-based trackers that aim to locate the most correlated region as the target region, the background information is processed in a discriminative framework. A classifier extracts the input features and then, the features are fed to the predictor to initialize the model and optimize the target prediction. The two key elements of DiMP tracker are the Target Center Regression (TCR) and Bounding Box Regression (BBR). The TCR output is a probabilistic map of the input image where the target region is coarsely warmed up in compared to the background pixels. The BBR output is the final bounding box prediction infers via a target conditional IoU-Net-based architecture~\citep{jiang_acquisition_2018} proposed in the ATOM~\citep{danelljan_atom_2019} tracker. 

\textbf{DiMP} \quad The Discriminative Model Prediction (DiMP) tracker~\citep{bhat_learning_2019} employs a discriminative approach that utilizes background information while also benefiting from the means of updating the target model. In contrast to Siamese-based trackers, which aim to identify the most correlated region as the target, DiMP processes background information within a discriminative framework. Initially, a classifier extracts input features, which are then utilized by a predictor to initialize the model and refine the target prediction. The DiMP tracker comprises two essential components: Target Center Regression (TCR) and Bounding Box Regression (BBR). TCR produces a probabilistic map of the input image, wherein the target region is delineated more prominently compared to the background pixels. The BBR component generates the final bounding box prediction through a target conditional IoU-Net-based architecture~\citep{jiang_acquisition_2018} proposed in the ATOM tracker~\citep{danelljan_atom_2019}.

\textbf{PrDiMP} \quad The Probabilistic Regression DiMP (PrDiMP)~\citep{danelljan_probabilistic_2020} aims to forecast the conditional probability density $p(y|x,\theta)$ of the target given the input frame. Training the PrDiMP tracker involves utilizing the Kullback-Leibler (KL) divergence between the predicted density $p(y|x,\theta)$ and the conditional ground truth $p(y|y_i)$. The conditional ground truth $p(y|y_i)$ is formulated to account for label noise and the ambiguity inherent in the regression task, addressing the uncertainty associated with annotations. PrDiMP builds upon the DiMP tracker~\citep{bhat_learning_2019} by incorporating probabilistic elements into both the Target Center Regression (TCR) and Bounding Box Regression (BBR) components.

%%The Probabilistic Regression DiMP (PrDiMP)~\citep{danelljan_probabilistic_2020} is aimed to predict the conditional probability density $p(y|x,\theta)$ of the target given the input frame. The KL divergence between the predicted density $p(y|x,\theta)$ and the conditional ground truth $p(y|y_i)$, is used to train PrDiMP tracker. The conditional ground truth $p(y|y_i)$ is defined to model the label noise and the ambiguity of the regression task. This feature is devised to encounter the uncertainty of the annotations themselves. The PrDiMP is based on the DiMP tracker~\citep{bhat_learning_2019} with the probabilistic TCR and the probabilistic BBR elements. 

% \textbf{SiamRPNDW} \quad

\subsection{Adversarial Attacks}

In our study, we examined four attack approaches against object trackers, as follows.

\paragraph{CSA} 
In attention-based Siamese trackers~\citep{li_siamrpn_2019}, the loss function aims to locate the hottest region in the image where the correlation of the target and that location is the highest among all other regions. Using two GANs, one for the template perturbation and the other for the search region perturbation, the CSA attack~\citep{yan_cooling-shrinking_2020} is developed to firstly cool the hot regions in the end of the network and then, shrink the object bounding box predicted by the tracker. The perturbation generators, GANs, are trained to predict the adversarial template and search regions. For the cooling loss term, the object candidates with smaller scores are eliminated and then, the remaining candidates are divided into positive $f_+$ and negative $f_-$ samples. The cooling term is computed as:
\begin{equation}
    L_\mathrm{cooling} = \frac{1}{N} \max(f_+ - f_-, m_c) \,,
\end{equation}
where $m_c$ is the margin for the classification labels. For the shrinking term, the less likely candidate similar to the cooling term is removed and then, the shrinking loss is calculated as
\begin{equation}
    L_\mathrm{shrinking} = \frac{1}{N} \max(R_w, m_w) + \frac{1}{N} \max(R_h, m_h) \,,
\end{equation}
where $m_w$ and $m_h$ are the margins for the width and height regression factors, respectively. In our experiments, we used the template and search GANs trained with the cooling-shrinking loss $L = L_\mathrm{cooling} + L_\mathrm{shrinking}$ to perturb the template and search regions for the trackers as a black-box attack. 

\paragraph{IoU} The IoU attack~\citep{jia_iou_2021} proposes a black-box setting attack to generate the adversarial frames based on the object motion with the purpose of decreasing the IoU between the predicted bounding box and the target. Two types of noises are added to achieve the final goal of the attack where the noise is bounded to a specific value for L1 norm. The IoU score $S_\mathrm{IoU}$ is defined as
\begin{equation}
    S_\mathrm{IoU} = \lambda S_\mathrm{spatial} + (1- \lambda) S_\mathrm{temporal} \,,
\end{equation}
where $S_\mathrm{spatial}$ is the actual intersection over union between the predicted bounding box before and after adding the noise, while the $S_\mathrm{temporal}$ is the intersection over union between current and previous predicted bounding box. Using this IoU score, the final predicted bounding box is aimed at decreasing both in temporal and spatial domains. The level of noise is controlled depending on the IoU score and a final upper bound limits the algorithm in adding noise to the frame.

\paragraph{SPARK} In SPARK~\citep{guo_spark_2020}, the classification and regression labels are manipulated to create the white-box attack. We employed the untargeted SPARK attack in our experiments. The SPARK aims to minimize the perturbation in a way that the final intersection over union of the current prediction and previous prediction is minimum. Furthermore, a new regularization term is used in SPARK as

\begin{equation}
    L_\mathrm{reg} = \lambda || \Gamma ||_{1,2}, \quad \Gamma = [\epsilon_{t-L}, ..., \epsilon_{t-1}, \epsilon_{t} ] \,,
\end{equation}

\noindent where $\Gamma$ represents the incremental perturbations of the last $L$ frames. The $\epsilon_t$ is computed as the $\epsilon_t = E_t - E_{t-1}$ in which $E_t$ is the current frame perturbation. The generated perturbation up to the last $L = 30$ frames is accumulated to create the adversarial search regions for the trackers. As a result, the computed perturbations are temporally and spatially sparse.

\paragraph{RTAA} Using RTAA~\citep{jia_robust_2020}, the classification and regression of object candidates are manipulated to generate the adversarial search regions. The classification labels are simply reversed and the regression labels are manipulated as follows:
\begin{align}
    & x_r^* = x_r + \delta_\mathrm{offset} \nonumber \\
    & y_r^* = y_r + \delta_\mathrm{offset} \nonumber \\
    & w_r^* = w_r \times \delta_\mathrm{scale} \nonumber \\
    & h_r^* = h_r \times \delta_\mathrm{scale} \,,
\end{align}

\noindent where $\delta_\mathrm{offset}$ and $\delta_\mathrm{scale}$ are the random distance and scale variations, respectively. The adversarial loss is computed as the difference of the original loss for true labels and the manipulated labels. Only the last frame perturbation is used from the past in the current step of the attack.

\subsection{Attack Setups}
\label{transfer}

In our study, we applied the attacks as they are proposed in the original works, with white-box attacks using the victim tracker's gradients, not transferring the attack from another tracker. For instance, SPARK~\citep{guo_spark_2020} is attacking trackers in a not-transferred white-box setting. As SPARK uses both classification and regression labels to generate the perturbation, it cannot be applied in a white-box setting for some trackers that are not providing them both -- e.g., \cite{cui_mixformer_2022, cai_robust_2023,bhat_learning_2019, danelljan_probabilistic_2020}. Table~\ref{tab:trans} specifies the applicable attacks on visual trackers investigated in our study. For instance, in MixFormer~\citep{cui_mixformer_2022}, the classification labels are fused by a score prediction module to infer one single score as the part of output, which is not compatible with SPARK. Also, in DiMP~\citep{bhat_learning_2019} and PrDiMP~\citep{danelljan_probabilistic_2020}, the regression labels for object candidates are not produced because these trackers predict the target bounding box directly. We have the same constraints with the RTAA attack~\citep{jia_robust_2020}, which also requires the classification and regression labels, and therefore, is not applicable as a white-box attack to the same trackers than SPARK. As for the CSA attack~\citep{yan_cooling-shrinking_2020}, since the template and search GANs are available, we are using these pre-trained networks and SiamRPN++~\citep{li_siamrpn_2019} tracker to generate the corresponding perturbations. CSA is therefore applied in the black-box setting similar to the IoU attack~\citep{jia_iou_2021}. However, CSA attack is also not applicable on DiMP and PrDiMP trackers, as they do not have template and search regions in the tracking process. This lead us to achieve our experiments over two white-box attacks, SPARK and RTAA, and two black-box attacks, CSA and IoU. 

\begin{table*}
    \centering 
    \caption{Applicability of adversarial attacks (rows) to tracking methods (columns). In this work, we evaluate: SPARK~\citep{guo_spark_2020}, RTAA~\citep{jia_robust_2020}, IoU~\citep{jia_iou_2021}, and CSA~\citep{yan_cooling-shrinking_2020}, against: ROMTrack~\citep{cai_robust_2023}, MixFormerM~\citep{cui_mixformer_2022}, TransT~\citep{chen_transformer_2021}, DiMP~\citep{bhat_learning_2019}, PrDiMP~\citep{danelljan_probabilistic_2020}, SiamRPN~\citep{li_high_2018}, and DaSiamRPN~\citep{zhu_distractor-aware_2018}. We identify combinations as either applicable (``A'') or not (``N/A'').}
    \label{tab:trans}
    \begin{tabular}{lccccccc}
    % {p{0.08\textwidth}p{0.1\textwidth}p{0.12\textwidth}p{0.08\textwidth}p{0.08\textwidth}p{0.08\textwidth}p{0.1\textwidth}p{0.12\textwidth}} \\ 
    \toprule
     & ROMTrack & MixFormerM & TransT & DiMP & PrDiMP & SiamRPN & DaSiamRPN \\
     \midrule
     SPARK(white-box) & N/A & N/A & A & N/A & N/A & A & A \\
     RTAA (white-box) &  N/A & N/A & A & N/A & N/A & A & A \\
     IoU (black-box) &  A & A & A & A & A & A & A \\
      CSA (black-box) & A & A & A & N/A & N/A & A & A \\
       \bottomrule
    \end{tabular}
\end{table*}

\section{Investigation}

We conducted an analysis to determine how sensitive transformer trackers are to perturbations generated by existing attack methods under various conditions. We compared the difference in performance between the tracker's ability to provide accurate bounding boxes and binary masks by measuring the percentage difference from their original performance on clean data. We evaluated the impact of adversarial attacks on transformer trackers in predicting the object bounding boxes by varying the perturbation levels. Finally, we assessed the performance of the IoU attack when the generated perturbations were bounded at different noise levels. We then discussed the observations we drew from these sets of experiments.

\subsection{Adversarial Attacks per Tracker Output}
\label{exp1}

In this section, we have applied adversarial attack techniques against the TransT-SEG~\citep{chen_high-performance_2023} and MixFormerM~\citep{cui_mixformer_2022} trackers and compared the results based on different tracking outputs. The objective of this experiment is to determine the difference in each tracking metric before and after the attack when one of the tracker's outputs (bounding box or binary mask) is measured. 

\textbf{Evaluation Protocol}\quad  We selected the VOT2022 Short-term dataset and protocol~\citep{kristan_tenth_2023} because of these three reasons. Unlike other datasets that use the one-pass evaluation protocol, the VOT2022-ST follows the anchor-based short-term protocol for the trackers evaluation. The metrics of VOT2022 baseline are obtained from the anchor-based short-term protocol which presents another view of the trackers' performances. Also, the VOT2022-ST provides the evaluation based on both object bounding boxes (STB) and binary masks (STS) which makes it a fair setup for our experiment. Therefore, the experiments for different attacks are achievable offline. Besides, the newest trackers are annually assessed by the VOT community and the most robust trackers are listed and announced on different sub-challenges. For instance, the MixFormerM tracker~\citep{cui_mixformer_2022} was among five top-ranked trackers for binary mask prediction and its other variant, MixFormerL won third place on the bounding box prediction sub-challenge. We have conducted a baseline experiment for the VOT2022~\citep{kristan_tenth_2023} short-term sub-challenge in two cases: object bounding box (STB) vs.\ object masks (STS) for target annotation and tracking. The Expected Average Overlap (EAO), accuracy and the anchor-based robustness metrics are calculated in this experiment. The EAO computes the expected value of the prediction overlaps with the ground truth. The accuracy measures the average of overlaps between tracker prediction and the ground truth over a successful tracking period. The robustness is computed as the length of a successful tracking period over the length of the video sequence. The successful period is a period of tracking in which the overlap between the prediction and ground truth is always greater than the pre-defined threshold. The baseline metrics are computed based on the anchor-based protocol introduced in VOT2020~\citep{kristan_eighth_2020}. In every video sequence evaluation under this protocol, the evaluation toolkit will reinitialize the tracker from the next anchor of the data to compute the anchor-based metrics wherever the tracking failure happens. For visualization usage, we employed some video sequences from DAVIS2016~\citep{perazzi_benchmark_2016} dataset. 

\textbf{Attacks Setting}\quad Four adversarial attacks are employed in this experiment, namely CSA~\citep{yan_cooling-shrinking_2020}, IoU~\citep{jia_iou_2021}, SPARK~\citep{guo_spark_2020}, and RTAA~\citep{jia_robust_2020}. However, not all of the attacks are applicable to both trackers. The SPARK and RTAA attacks manipulate the object candidates list, which includes classification labels and/or regression targets. If the tracker does not infer the candidates in the output, these attacks cannot be applied on, such as MixFormerM~\citep{cui_mixformer_2022} that only outputs the predicted bounding box. In this experiment, we generated the perturbations using SPARK and RTAA attacks, i.e. white-box attacks, against trackers. However, the perturbation of CSA~\citep{yan_cooling-shrinking_2020} is created by two GANs and passing the image into the SiamseRPN++~\citep{li_siamrpn_2019} tracker to generate the adversarial loss depending on the SiamseRPN++ loss. Therefore, the CSA is a transferred black-box attack for both TransT-SEG~\citep{chen_high-performance_2023} and MixFormer~\citep{cui_mixformer_2022} trackers. The IoU method~\citep{jia_iou_2021} is also a black-box approach that can perturb the whole frame using the tracker prediction for several times. 

%or ROMTrack~\citep{cai_robust_2023}.

%In addition to the reproducibility of existing attack ideas, we will check if these ideas are generalizable to the transformer-based trackers.

\textbf{Results}\quad The following is a summary of the results obtained from an experiment conducted on the VOT2022~\citep{kristan_tenth_2023} dataset using the TransT-SEG tracker~\citep{chen_high-performance_2023} after adversarial attacks. The results are shown in Table~\ref{tab:votT} for both STB and STS cases. The most powerful attack against TransT-SEG~\citep{chen_high-performance_2023} in all three metrics was found to be SPARK~\citep{guo_spark_2020}. According to the accuracy scores of Table~\ref{tab:votT}, the object binary mask was more affected by the adversarial attacks than the object bounding box. However, the difference of drop percentages in assessing the predicted bounding boxes and binary masks are negligible for the EAO and Robustness metrics of TransT-SEG tracker.

%The evaluation metrics revealed that the object binary mask was more affected by the adversarial attacks than the object bounding box.

\begin{table*}
    \centering 
    \caption{Evaluation results of the TransT-SEG~\citep{chen_high-performance_2023} tracker attacked by different methods on the VOT2022~\citep{kristan_tenth_2023} Short-Term (ST) dataset and protocol for two stacks: bounding box prediction via bounding box annotations (STB) and binary mask prediction via binary mask annotations (STS). The ``Clean'' values are the original tracker performance without applying any attack.}
    \label{tab:votT}
    \begin{tabular}{p{0.06\textwidth}p{0.08\textwidth}p{0.065\textwidth}p{0.065\textwidth}p{0.065\textwidth}p{0.065\textwidth}p{0.065\textwidth}p{0.065\textwidth}p{0.065\textwidth}p{0.065\textwidth}p{0.065\textwidth}p{0.065\textwidth}} \\ 
    \toprule
    & & \multicolumn{3}{c}{EAO} & \multicolumn{3}{c}{Accuracy}  & \multicolumn{3}{c}{Robustness}  \\
    \midrule
     Stack & Method & Clean & Attack & Drop & Clean & Attack & Drop &  Clean & Attack & Drop  \\
    \midrule 
    \multirow{4}{*}{STB} 
       & CSA & 0.299 & 0.285 & 4.68$\%$ & 0.472 & 0.477 & -1.06$\%$ & 0.772 & 0.744 & 3.63$\%$  \\
       & IoU & 0.299 & 0.231 & 22.74$\%$ & 0.472 & 0.495 & -4.87$\%$ & 0.772 & 0.569 & 26.29$\%$  \\
       & RTAA & 0.299 & 0.058 & 83.28$\%$ & 0.472 & 0.431 & 8.69$\%$ & 0.772 & 0.157 & 79.66$\%$  \\
       & SPARK & 0.299 & 0.012 & 95.99$\%$ & 0.472 & 0.244 & 48.30$\%$ & 0.772 & 0.051 & 93.39$\%$  \\
    \midrule
    \multirow{4}{*}{STS} 
       & CSA & 0.500 & 0.458 & 8.40$\%$ & 0.749 & 0.736 & 1.73$\%$ & 0.815 & 0.779 & 4.42$\%$  \\
       & IoU & 0.500 & 0.334 & 33.20$\%$ & 0.749 & 0.710 & 5.21$\%$ & 0.815 & 0.588 & 27.85$\%$  \\
       & RTAA & 0.500 & 0.067 & 86.60$\%$ & 0.749 & 0.533 & 28.84$\%$ & 0.815 & 0.146 & 82.08$\%$  \\
       & SPARK & 0.500 & 0.011 & 97.80$\%$ & 0.749 & 0.266 & 64.48$\%$ & 0.815 & 0.042 & 94.84$\%$  \\
       \bottomrule
    \end{tabular}
\end{table*}

It was observed that the CSA~\citep{yan_cooling-shrinking_2020} attacks poorly degraded the outputs in evaluation metrics, except for the accuracy of the STB case. However, it was surprising to note that there was a negative difference in the accuracy metric after the CSA and IoU attack. Specifically, in the STB case, after the adversarial attacks, the accuracy of TransT-SEG~\citep{chen_high-performance_2023} decreased by $-1.06\%$ for the CSA attack and $-4.87\%$ for the IoU attack. This improvement was also observed for MixFormer~\citep{cui_mixformer_2022} in Table~\ref{tab:votM} for EAO and robustness metrics after the CSA attack in the STB case and accuracy after the CSA in the STS case of the experiments. The binary masks predicted by the MixFormerM tracker after the IoU attack for all computed metrics dropped greater than the metrics computed for the predicted bounding boxes. Nonetheless, following the CSA attack, the drop percentages of metrics for MixFormerM outputs were smaller and more consistent compared to those resulting from the IoU attack. Comparing the IoU attack results on both Tables~\ref{tab:votT} and~\ref{tab:votM} for bounding box evaluation (STB), the MixFormerM shows greater adversarial robustness with $14.74\%$ drop percentage from original score whereas the robustness of TransT dropped by $26.29\%$ after the same attack.

According to Table~\ref{tab:votM}, the most powerful attack against MixFormerM~\citep{cui_mixformer_2022} is the IoU attack~\citep{jia_iou_2021}. Even after the IoU attack~\citep{jia_iou_2021}, the adversarial perturbation slightly improves accuracy in assessing object binary masks (STS) and EAO in assessing predicted bounding boxes (STB). The EAO metric in the evaluation of the bounding box and binary mask is the most affected metric with $18.81\%$ and $39.05\%$ drop percentages after IoU attack. When compared to the corresponding metric for TransT-SEG~\citep{chen_high-performance_2023} as shown in Table~\ref{tab:votT}, the IoU attack had a greater damage on binary mask creation for MixFormerM than for TransT-SEG per EAO and accuracy metrics. However, this result was reversed for the same metrics when the object bounding box was evaluated in both trackers, Tables \ref{tab:votT} and~\ref{tab:votM} (STB). For this point, it is important to mention that tracking and segmentation is performed by two different networks in MixFormerM~\citep{cui_mixformer_2022}. Therefore, the object bounding box evaluation is the assessment of the tracker's network while the binary mask evaluation is the assessment of the segmentation network~\citep{yan_alpha-refine_2021}. In contrast, the TransT-SEG~\citep{chen_high-performance_2023} tracker performs both tracking and segmentation by a single transformer tracker.

\begin{table*}
    \centering 
    \caption{Evaluation results of the MixFormerM~\citep{cui_mixformer_2022} tracker attacked by different methods on the VOT2022~\citep{kristan_tenth_2023} Short-Term (ST) dataset and protocol for two stacks: bounding box prediction via bounding box annotations (STB), binary mask prediction via binary mask annotations (STS). The ''Clean'' values are the original tracker performance without applying any attack.}
    \label{tab:votM}
    \begin{tabular}{p{0.06\textwidth}p{0.08\textwidth}p{0.065\textwidth}p{0.065\textwidth}p{0.065\textwidth}p{0.065\textwidth}p{0.065\textwidth}p{0.065\textwidth}p{0.065\textwidth}p{0.065\textwidth}p{0.065\textwidth}p{0.065\textwidth}} \\ 
    \toprule
    & & \multicolumn{3}{c}{EAO} & \multicolumn{3}{c}{Accuracy}  & \multicolumn{3}{c}{Robustness}  \\
    \midrule
     Stack & Method & Clean & Attack & Drop & Clean & Attack & Drop &  Clean & Attack & Drop  \\
    \midrule 
    \multirow{2}{*}{STB} 
       & CSA & 0.303 & 0.308 & -1.65$\%$ & 0.479 & 0.478 & 0.21$\%$ & 0.780 & 0.791 & -1.41$\%$  \\
       & IoU & 0.303 & 0.246 & 18.81$\%$ & 0.479 & 0.458 & 4.38$\%$ & 0.780 & 0.665 & 14.74$\%$  \\
    \midrule
    \multirow{2}{*}{STS} 
       & CSA & 0.589 & 0.562 & 4.58$\%$ & 0.798 & 0.803 & -0.63$\%$ & 0.880 & 0.857 & 2.61$\%$  \\
       & IoU & 0.589 & 0.359 & 39.05$\%$ & 0.798 & 0.660 & 17.30$\%$ & 0.880 & 0.677 & 23.07$\%$  \\
       \bottomrule
    \end{tabular}
\end{table*}

%Figure~\ref{fig:exp1} demonstrates a sample frame perturbed by different attacks for MixFormerM and TransT-SEG trackers and the output results are depicted on the video frame. As the quantitative results indicated in Tables~\ref{tab:votT} and \ref{tab:votM}, the white-box attacks (SPARK and RTAA) have harmed the binary mask and bounding box more than the black-box attacks (IoU and CSA). 

\begin{figure*}
 \centering
 \centerline{\includegraphics[width=\textwidth]{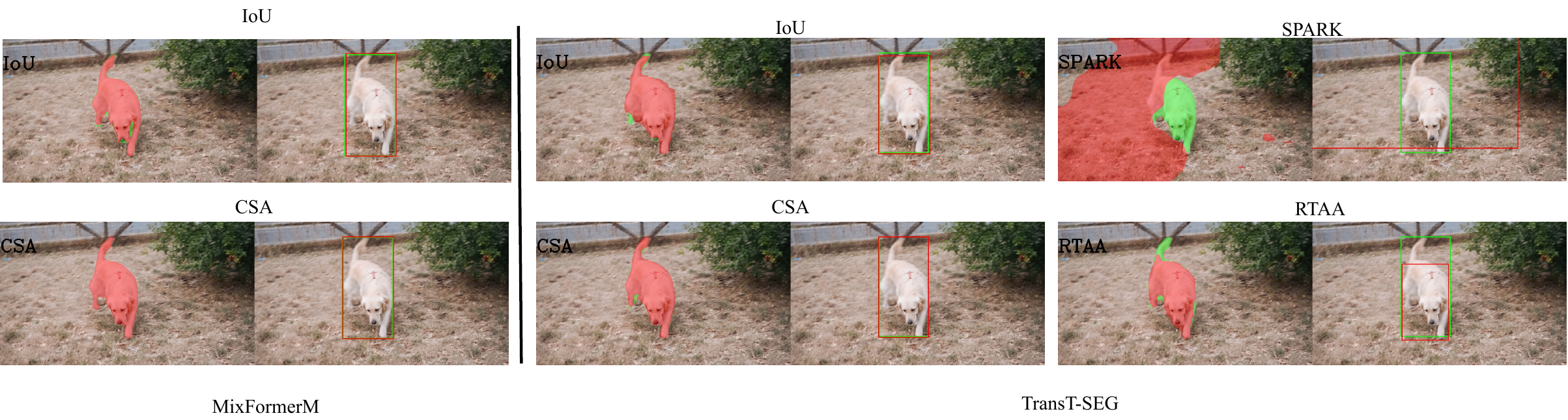}}
 \caption{Mask vs. bounding box predictions as the output of transformer trackers, MixFormerM~\citep{cui_mixformer_2022} and TransT-SEG~\citep{chen_high-performance_2023}, while the adversarial attacks applied to perturb the input frame/search region. The TransT-SEG tracker's outputs harmed by the white-box methods, SPARK~\citep{guo_spark_2020} and RTAA~\citep{jia_robust_2020}, more than black-box attacks, IoU~\citep{jia_iou_2021} and CSA~\citep{yan_cooling-shrinking_2020}. The green mask/bounding box represents the object tracker's performance while the red mask/bound box belongs to the tracker's performance after each attack.}
 \label{fig:exp1}
\end{figure*}

Figure~\ref{fig:exp1} demonstrates the results of different attacks on MixFormerM and TransT-SEG trackers in terms of the object bounding boxes and binary masks. The original outputs, bounding boxes and masks, are depicted by Green color, while the results after the attacks are exhibited with Red color. As the quantitative results indicated in Tables~\ref{tab:votT} and \ref{tab:votM}, the white-box attacks (SPARK and RTAA) have harmed the binary mask and bounding box more than the black-box attacks (IoU and CSA).

\subsection{Adversarial Attacks per Perturbation Level}
\label{exp2}

We test the effect of the perturbation levels on adversarial attack performance against transformer trackers. In white-box attacks such as SPARK~\citep{guo_spark_2020} and RTAA~\citep{jia_robust_2020}, the generated perturbation is used to update the frame patch in each attack step. The overview of pseudocode of these two attacks is presented in Algorithms~\ref{alg:RTAA} and \ref{alg:SPARK} where the $\phi_{\epsilon}$ indicates the operation of clipping the frame patch to the $\epsilon$-ball. The $\alpha$ is the applied norm of gradients, and $I$ is the search region. Although there are several differences in both settings, there is one similar step to generate the adversarial region from the input image gradient and previous perturbation(s). Line 4 of the RTAA pseudocode and line 6 of the SPARK pseudocode change the image pixels based on the computed gradients. By adjusting the $\epsilon$-ball, the performance of attacks is evaluated to demonstrate the power of each adversarial idea for object trackers. The plus or minus sign of the gradient sign corresponds to the decoupling direction and norm research in the gradient-based adversarial attack~\citep{rony_decoupling_2019}. For instance, SPARK~\citep{guo_spark_2020} uses minus, while RTAA~\citep{jia_robust_2020} sums up the sign of gradients with image values. Furthermore, note that in the original papers of SPARK~\citep{guo_spark_2020} and RTAA~\citep{jia_robust_2020}, the attack parameters may have different names than those we used in this paper. Our goal was to unify their codes and approach into principle steps that make comparison more accessible for the audience. An important aspect of the SPARK algorithm, mentioned in \citep{guo_spark_2020}, is its regularization term. This feature is convenient for maintaining sparse and imperceptible perturbations~\citep{guo_spark_2020}. The regularization term involves adding the $L_{2,1}$ Norm of previous perturbations to the adversarial loss, which helps generate sparse and imperceptible noises. We generated examples of SPARK perturbation~\citep{guo_spark_2020} versus RTAA~\citep{jia_robust_2020} perturbations to verify this claim.

\begin{algorithm}[tb]
\caption{RTAA~\citep{jia_robust_2020} algorithm as the adversarial attack for object trackers}\label{alg:RTAA}
\begin{algorithmic}[1]
\State $\mathcal{P} \gets \mathcal{P}(t-1)$ \Comment{Initialize with perturbation map of previous frame}
\State $I^\mathrm{adv} \gets I$ \Comment{Initialize with clean current frame}
\For {$i = 1,\ldots,i^\textrm{max}$} 
\State $I^\mathrm{adv} \gets  I^\mathrm{adv} + \phi^{\epsilon} \left (\mathcal{P} +\alpha \sign(\nabla_{I^\mathrm{adv}} \mathcal{L}) \right)$ \Comment{Application of adversarial gradient descent}
\State $I^\mathrm{adv} \gets \max(0, \min(I^\mathrm{adv},255))$ \Comment{Clamp image values in $[0,\,255]$}
\State $\mathcal{P} \gets I^\mathrm{adv} - I$\Comment{Update perturbation map}
\EndFor 
\State \textbf{Return} $I^\mathrm{adv}, \mathcal{P}$ \Comment{Return adversarial image and corresponding perturbation map}
\end{algorithmic} 
\end{algorithm}

\begin{algorithm}[tb]
\caption{SPARK~\citep{guo_spark_2020} algorithm as the adversarial attack for object trackers}\label{alg:SPARK}
\begin{algorithmic}[1]
\State $\mathcal{P} \gets \mathcal{P}(t-1)$ \Comment{Initialize with perturbation map of previous frame}
\State $\mathcal{S} \gets \sum_{i=1}^K \mathcal{P}(t-i)$ \Comment{Sum of perturbation maps of last $K$ frames}
\State $I^\textrm{adv} \gets I$ \Comment{Initialize with clean current frame image}
\For {$i=1,\ldots,i^\textrm{max}$} 
\State $I' \gets I^\textrm{adv}$\Comment{Get a copy of current adversarial image}
\State $I^\textrm{adv} \gets  I' + \phi^{\epsilon} \left( \mathcal{P} - \alpha \sign(\nabla_{I'} \mathcal{L} ) \right) + \mathcal{S}$ \Comment{Application of adversarial gradient descent}
\State $I^\textrm{adv} \gets \max(0, \min(I^\textrm{adv}, 255))$ \Comment{Clamp image values in $[0,\,255]$}
\State $\mathcal{P} \gets I^\textrm{adv} - I' - \mathcal{S} $\Comment{Update perturbation map} \smallskip
\EndFor
\State $\mathcal{S} \gets \sum_{i=1}^K \mathcal{P}(t-i)$ \Comment{Update the sum of perturbation maps} \smallskip
\State $I^\mathrm{adv} \gets I + \mathcal{S}$ \Comment{Generate the current adversarial frame}\smallskip
\State \textbf{Return} $I^\textrm{adv}, \mathcal{P}$\Comment{Return adversarial image and corresponding perturbation map} 
\end{algorithmic} 
\end{algorithm}

\textbf{Evaluation Protocol}\quad The test sets of the experiments on the perturbation level changes are the UAV123 dataset~\citep{mueller_benchmark_2016} and VOTST2022~\citep{kristan_tenth_2023}. The UAV123 dataset comprises 123 video sequences with natural and synthetic frames in which an object appears and disappears from the frame captured by a moving camera. We calculate success and precision rates across various thresholds under the One Pass Evaluation (OPE) protocol. In this setup, the object tracker is initialized using the first frame and the corresponding bounding box. Subsequently, the tracker is evaluated for each frame's prediction for the rest of the video sequence. Precision is measured by calculating the distance between the center of the ground truth's bounding box and the predicted bounding box. The precision plot shows the percentage of bounding boxes that fall within a given threshold distance. The success rate is computed based on the Intersection over Union (IoU) between the ground truth and predicted bounding boxes. The success plot is generated by considering different thresholds over IoU and computing the percentage of bounding boxes that pass the given threshold. Furthermore, we computed the L1 norm and structural similarity (SSIM)~\citep{wang_image_2004} as the measurements of sparsity and imperceptibility of the generated perturbations per attack. We chose some frames of the VOT2022ST~\citep{kristan_tenth_2023} dataset to visualize these metrics per frame.

\textbf{Attacks Setting}\quad The SPARK~\citep{guo_spark_2020} and RTAA~\citep{jia_robust_2020} approaches applied on the TransT tracker~\citep{chen_transformer_2021} are assessed in this experiment using the OPE protocol. We chose the TransT tracker which is a pioneer on transformer trackers to observe the attack performance change on the perturbation levels. Both attacks generate the perturbed search region over a fixed number of iterations (10). While the step size $\alpha$ for the gradient's update is $1$ for RTAA and $0.3$ for SPARK. We used five levels of perturbation $\epsilon \in \{ 2.55, 5.1, 10.2, 20.4, 40.8 \}$ to compare its effects on the TransT~\citep{chen_transformer_2021} performance on UAV123~\citep{mueller_benchmark_2016} and VOT2022ST~\citep{kristan_tenth_2023} datasets. The $\epsilon$'s are selected as a set of coefficients $\{ 0.01, 0.02, 0.04, 0.08, 0.16 \}$ of the maximum pixel value $255$ in an RGB image. It is worth mentioning that the $\epsilon$ for both attacks are set to $10$ in their original settings. Therefore, the original performance of each attack is very close to the $\epsilon_3 = 10.2$ perturbation level.

\begin{figure*}
     \centering
     \begin{subfigure}[b]{0.23\textwidth}
         \centering         \centerline{\includegraphics[width=\textwidth]{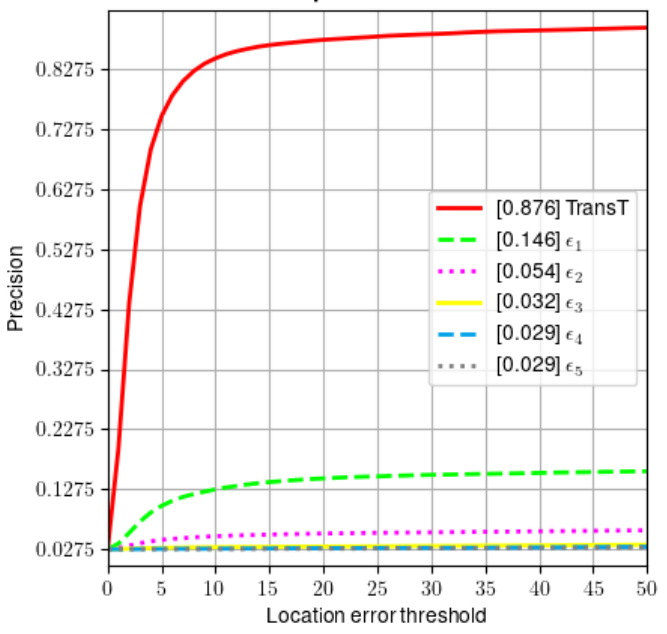}}
         \caption{}
     \end{subfigure}
     \hfill
     \begin{subfigure}[b]{0.23\textwidth}
         \centerline{\includegraphics[width=\textwidth]{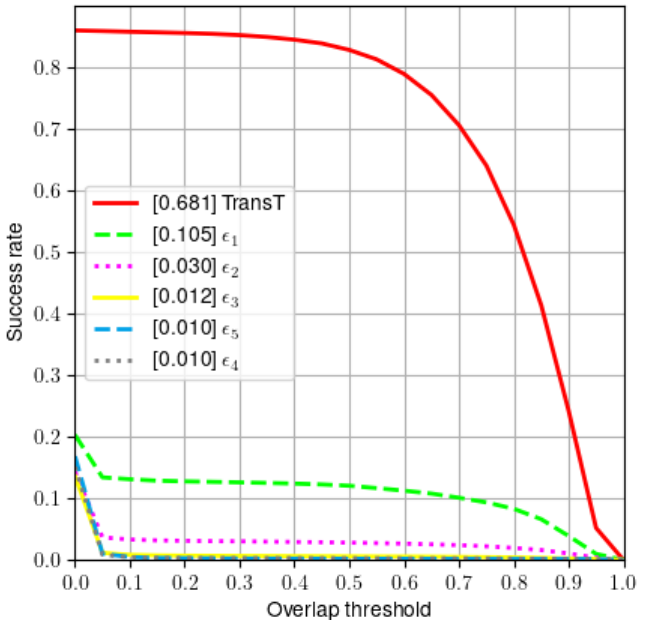}}
         \caption{}
     \end{subfigure}
     \begin{subfigure}[b]{0.23\textwidth}
         \centering
         \centerline{\includegraphics[width=\textwidth]{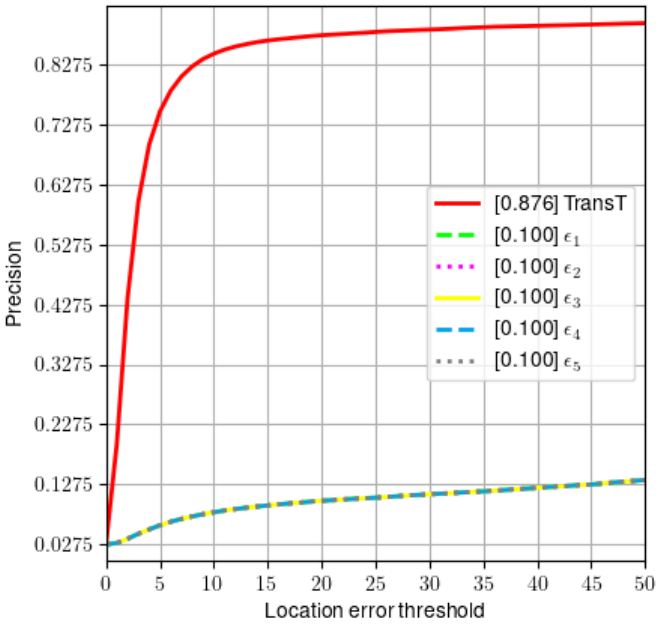}}
         \caption{}
     \end{subfigure}
     \hfill
     \begin{subfigure}[b]{0.23\textwidth}
         \centerline{\includegraphics[width=\textwidth]{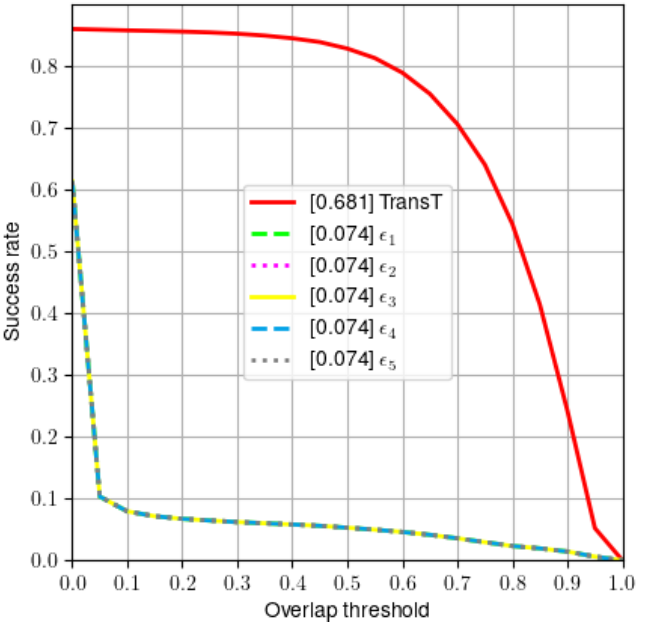}}
         \caption{}
     \end{subfigure}
     \caption{The precision and success plots related to the TransT~\citep{chen_transformer_2021} performance after RTAA~\citep{jia_robust_2020} (a, b) and SPARK~\citep{guo_spark_2020} (c,d) attack under different levels of noise on UAV123~\citep{mueller_benchmark_2016} dataset. The average score for each metric is shown in the legend of the plots. The 'red' plot is the original TransT performance without any attack applied on the tracker. The $e$'s are corresponded to $\epsilon$'s in our experiment, changing from $e_1 = 2.55$ to $e_5 = 40.8$ to assess the TransT performance after the white-box attacks under various perturbation levels. The SPARK performances per perturbation level shifts did not change on UAV123dataset as one can observe the SPARK curves are overlapped. } 
     \label{fig:pert}
\end{figure*}

\textbf{Results}\quad Figure~\ref{fig:pert} shows the performance of TransT~\citep{chen_transformer_2021} under RTAA~\citep{jia_robust_2020} and SPARK~\citep{guo_spark_2020} attacks with different perturbation levels. The red curve indicates the clean performance of the tracker before applying any attacks. The other perturbation levels are demonstrated with different colors. Unlike classification networks with transformer backbones~\citep{shao_adversarial_2022}, the transformer tracker performances after the RTAA attack~\citep{jia_robust_2020} using different $\epsilon$'s are minimally different but not after SPARK attacks~\citep{guo_spark_2020}. Adversarial perturbation methods against trackers use the previous perturbation to be added to the current frame. This setting may remove the sensitivity of the attack methods in the perturbation levels. In the RTAA attack~\citep{jia_robust_2020}, only one last perturbation is added to the current frame. In contrast, the SPARK~\citep{guo_spark_2020} uses the previous perturbations in each time step for the last $K=30$ frames, which reduces the sensitivity of the output to small changes in the inputs.  For perturbation levels $\{\epsilon_3, \epsilon_4, \epsilon_5\}$, RTAA's performance~\citep{jia_robust_2020} remains the same, whereas using smaller levels affects its performance. It is noteworthy that RTAA~\citep{jia_robust_2020} outperforms SPARK~\citep{guo_spark_2020} on UAV datasets~\citep{mueller_benchmark_2016} in almost every perturbation level except for the most minor level $\epsilon_1 = 0.01$ in which SPARK is the stronger attack.

\begin{figure*}
 \centering
 \centerline{\includegraphics[width=\textwidth]{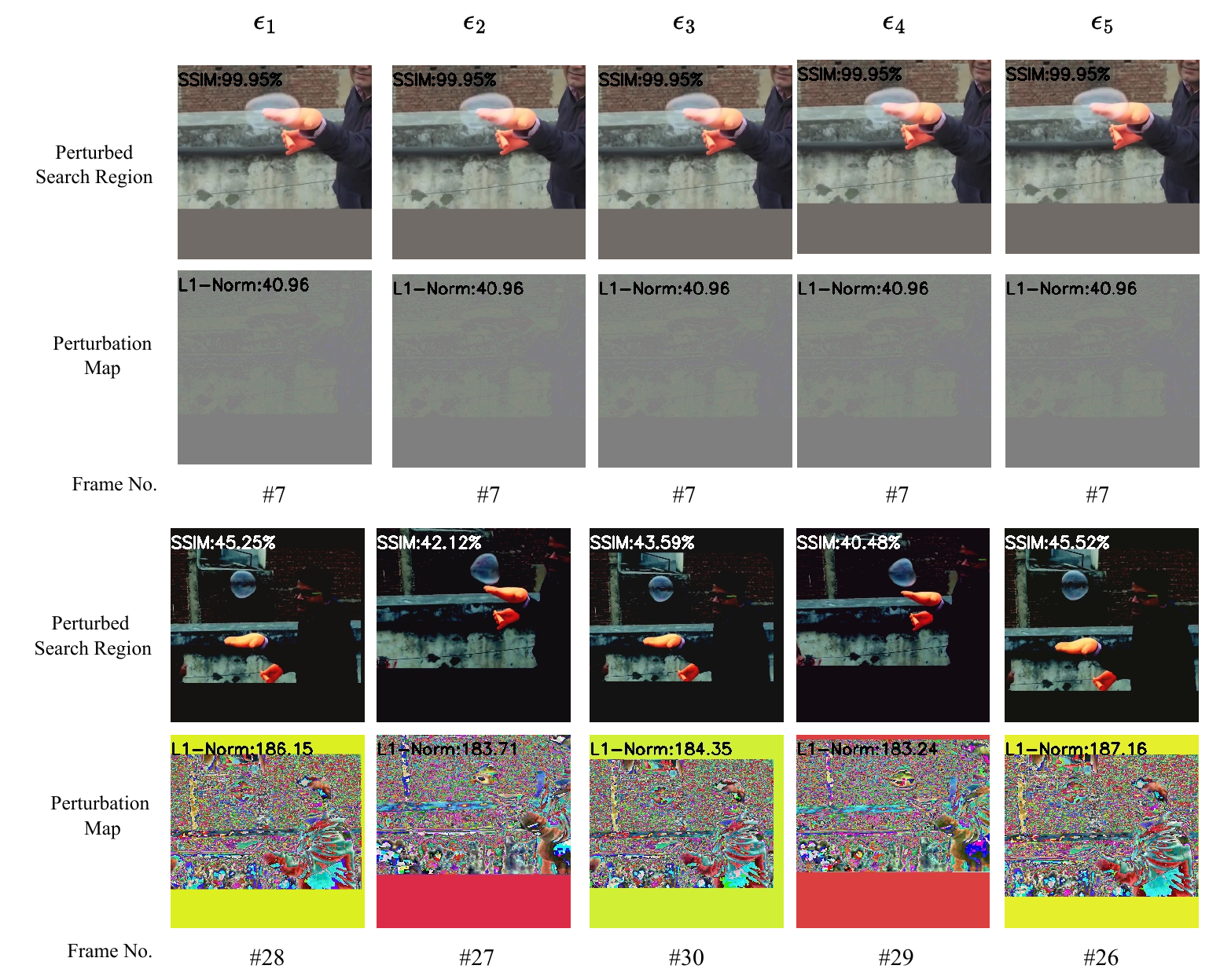}}
 \caption{The search regions related to the ``bubble'' sequence in the VOT2022ST dataset~\citep{kristan_tenth_2023} after applying SPARK~\citep{guo_spark_2020} attack on TransT~\citep{chen_transformer_2021} tracker. The perturbed search region is labeled with the SSIM~\citep{wang_image_2004} measured between search regions before and after the attack. The perturbation maps, following the work of~\citep{yan_cooling-shrinking_2020}, are created to demonstrate the added noise in colors. The L1 norm for perturbation maps are calculated to show the perturbation density/sparsity.}
 \label{fig:spark}
\end{figure*}

In the main paper of SPARK~\citep{guo_spark_2020}, it has been mentioned that the technique generates a temporally sparse and imperceptible noise. Figure~\ref{fig:spark} displays various examples of perturbed search regions and perturbation maps produced by the TransT tracker after applying the SPARK attack. Upon applying the attack, we noted that some frames, like frame number "7" of the 'bubble' sequence and the first two rows of Figure~\ref{fig:spark}, generated search regions and perturbation maps with fixed values for imperceptibility (SSIM metric) and sparsity (L1 norm). Even by increasing the perturbation level, some frames retained the same level of imperceptibility and sparsity. However, there were also instances of super-perturbed search regions per video sequence, where the noise was noticeable, and the L1 norm had a high value, as shown in the last two rows of Figure~\ref{fig:spark}. We consider a perturbed search region super-perturbed when the imperceptibility of the region is lower than $50\%$. The SPARK algorithm generates the most imperceptible noise with a constant and high SSIM value of $99.95\%$ and sparse noise with L1 norm of $40.96$ in all of the perturbation levels given the same frame. This fixed number of L1 norm is also the result of the regularization term discussed in the SPARK paper~\citep{guo_spark_2020}. This stability of SSIM and L1 norm have been repeated for many frames of the ``bubble'' sequence for SPARK attack~\citep{guo_spark_2020} while in some frames, the imperceptibility and sparsity are not stable per perturbation levels.

In Figure~\ref{fig:spark}, we indicate some super-perturbed regions with their perturbation maps per perturbation level. Interestingly, as we increase the perturbation levels, the number of super-perturbed regions also increases. In the attack settings, the perturbation of previous frames considered in the loss function is erased every $30$ frames for RTAA~\citep{jia_robust_2020} and SPARK~\citep{guo_spark_2020} algorithms. Table~\ref{tab:perts} provides information about the number of highly perturbed frames during a video sequence and the average imperceptibility (SSIM) and sparsity (L1 norm) scores. With higher levels of perturbations, more frames become highly perturbed, resulting in a greater L1 norm of the perturbations. Furthermore, in the lower perturbations levels, the highly perturbed search regions generate more perceptible noise, i.e. the imperceptibility of generated perturbations have grown by boosting the perturbation level.

%For many frames such as frame $\#7$ the search regions are generated with the fixed imperceptibility and sparsity. In contrast, there are some super-perturbed search regions per sequence which grows in the number with boosting the perturbation level.

\begin{table*}
    \centering 
    \caption{The perturbation levels versus number of highly perturbed search regions generated by the SPARK algorithm~\citep{guo_spark_2020} applied on TransT~\citep{chen_transformer_2021} tracker. The SSIM and L1 norm are computed as the average number of highly perturbed regions on the ``bubble'' sequence of VOT2022~\citep{kristan_tenth_2023} dataset. The "No. of frames" is the number of super-perturbed frames in which the SSIM value is below than $50\%$.}
    \label{tab:perts}
    \begin{tabular}{cccc}
    % {p{0.2\textwidth}p{0.2\textwidth}p{0.2\textwidth}p{0.2\textwidth}} \\ 
    \toprule
     $\epsilon$  & No. of frames & SSIM & L1 norm \\
    \midrule
     $2.55$ & $7$ & $36.86$ & $176.04$ \\
     $5.1$ & $ 7$ & $40.96$ & $181.86$\\
     $10.2$ & $ 13$ & $41.08$ & $181.33$\\
     $20.4$ & $13$ & $41.97$& $182.53$\\
     $40.8$ & $14$ & $42.53$& $183.98$ \\
       \bottomrule
    \end{tabular}
\end{table*}

\begin{figure*}
 \centering
 \centerline{\includegraphics[width=\textwidth]{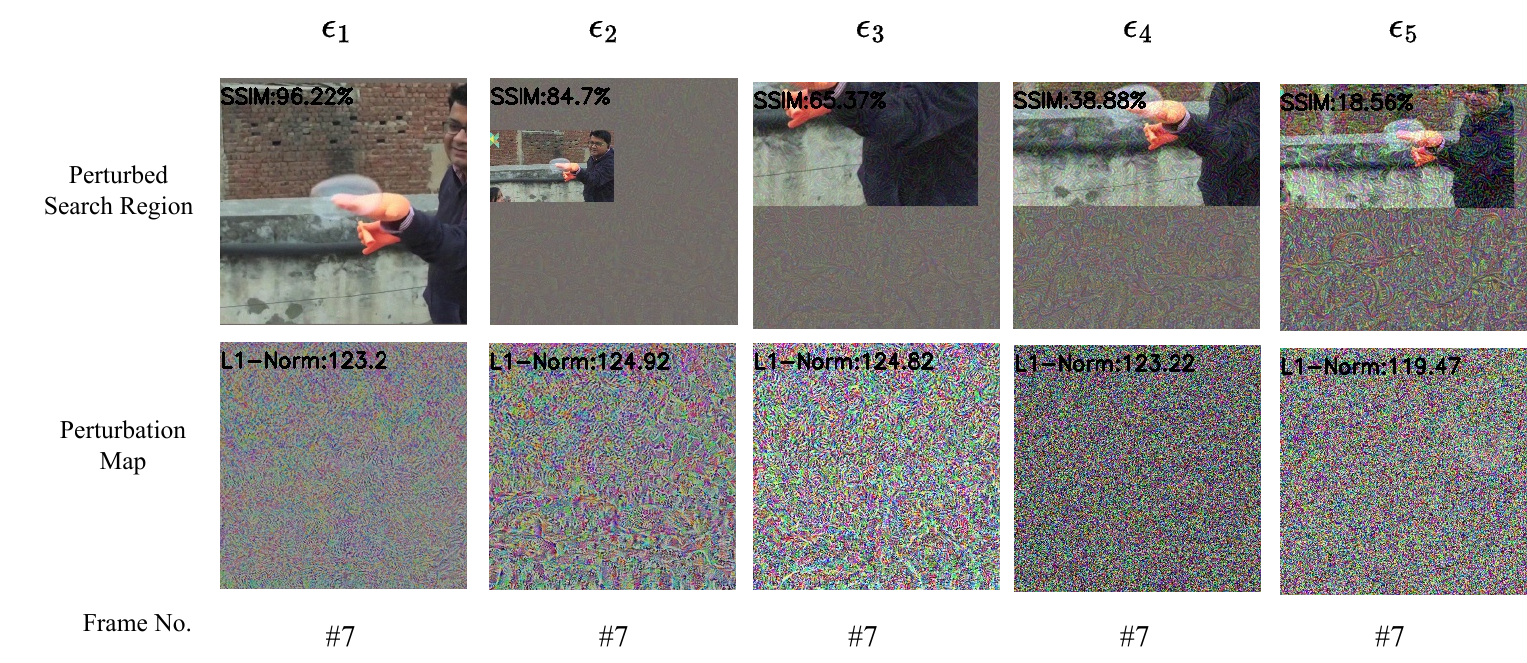}}
 \caption{The search regions related to the ``bubble'' sequence in the VOT2022ST dataset~\citep{kristan_tenth_2023} after applying RTAA~\citep{jia_robust_2020} attack on TransT~\citep{chen_transformer_2021} tracker. The perturbed search region is labeled with the SSIM~\citep{wang_image_2004} measured between search regions before and after the attack. The perturbation maps, following the work of~\citep{yan_cooling-shrinking_2020}, are created to demonstrate the added noise in colors. The L1 norm for perturbation maps are calculated to show the perturbation density (i.e. sparsity).}
 \label{fig:rtaa}
\end{figure*}

For the RTAA~\citep{jia_robust_2020} attack applied on the TransT~\citep{chen_transformer_2021} tracker, whenever the perturbation level boosts the imperceptibility and sparsity declines. Figure~\ref{fig:rtaa} demonstrates the result of applying RTAA against TransT tracker for the same frame $\#7$ of the 'bubble' sequence. The RTAA attack perturbs search regions with higher SSIM values at the lowest $\epsilon$ level, i.e. the first level $\epsilon=2.55$. By increasing the perturbation levels, the perceptibly of the RTAA perturbation has been increased while the sparsity changes are small.

\subsection{Adversarial Attack per Upper-Bound}
\label{exp3}

We conducted an experiment to test the vulnerability of the IoU attack~\citep{jia_iou_2021} to noise bounding, using different upper bounds. The IoU method~\citep{jia_iou_2021} is a black-box attack that misleads trackers by adding various noises to the frame using object bounding boxes. The essential steps of the IoU attack~\citep{jia_iou_2021} involve creating two levels of noise perturbations: orthogonal and normal direction noises. Our study aims to manipulate the attack settings in the second part of perturbation generation, which is in the normal direction.

\textbf{Dataset and Protocol}\quad The performances of the IoU attack are assessed against ROMTrack~\citep{cai_robust_2023} on UAV123~\citep{mueller_benchmark_2016} dataset using the OPE protocol. The success rate, precision rate, and normalized precision rate are computed to compare the results. For this experiment, we report the average of the success rate called Area Under Curve (AUC), as well as the average precision and norm precision on the thresholds. 

\textbf{Attack Setting}\quad The IoU method~\citep{jia_iou_2021} is a black-box attack on object trackers. It adds two types of noise to the frames: one in the tangential direction and the other in the normal direction. In the original setting, there was no limit on the number of times that noise could be added in the normal direction. This noise was limited only by the upper bound $\zeta$ and the $S_{IoU}$ value in the original setting. The $S_{IoU}$ value is the weighted average of two computed IoU values: 1) the IoU of the noisy frame prediction and the first initialized frame $I^{adv}_1$ in the attack algorithm, and 2) the IoU of the noisy frame prediction and the last frame bounding box in the tracking loop. In our experiment, we set a limit of $10$ steps in the algorithm's last loop to reduce the processing time, especially for the larger upper bound $\zeta$ values. We tested the IoU attack under three upper bounds: $\zeta \in \{ 8000, 10000, 12000 \}$. The middle value of $\zeta = 10000$ corresponds to the original setting of the IoU attack~\citep{jia_iou_2021}.

\begin{figure*}
 \centering
 \centerline{\includegraphics[width=\textwidth]{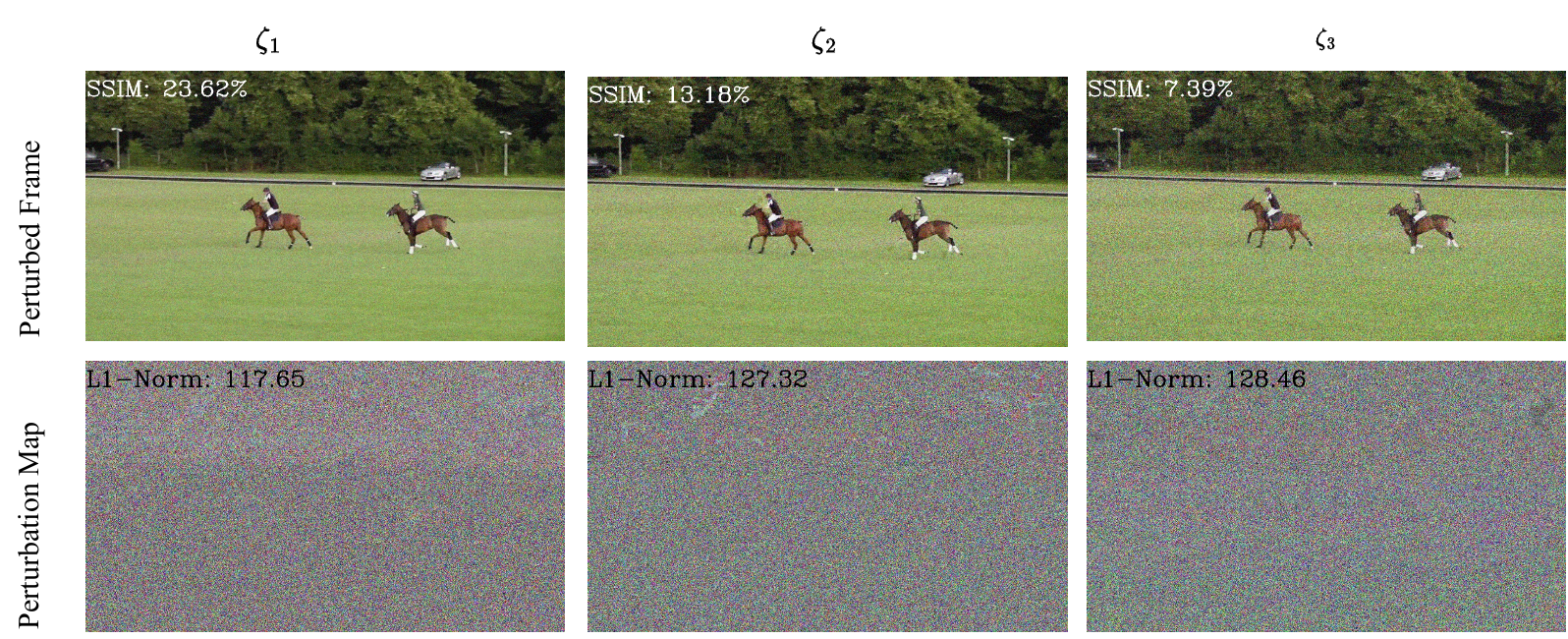}}
 \caption{The perturbed frames and perturbation maps generated by the IoU method~\citep{jia_iou_2021} against ROMTrack~\citep{cai_robust_2023} using three upper bounds of $\zeta \in \{8k, 10k, 12k \}$. The imperceptibility and L1 norm of the generated perturbations are shown in the frames representing the noise imperceptibility and sparsity of perturbation maps.}
 \label{fig:iou}
\end{figure*}

\textbf{Results}\quad The images in Figure~\ref{fig:iou} display the results of the IoU attack~\citep{jia_iou_2021} against ROMTrack~\citep{cai_robust_2023} under various upper bounds for a single frame. The L1 norm of the perturbation has increased as the upper bounds were raised. Additionally, the imperceptibility, measured by the SSIM values, decreased as the perturbations became more severe. Since the IoU attack starts by generating some random noise, it is highly dependent on the initialization points. For some cases, the algorithm did not process a single video sequence even after $48$ hours. One solution that worked for proceeding was to stop the processing without saving any results about the current sequence to restart the evaluation. After re-initialization, the attack began from another random point (noise) and it proceeded to the next sequence in less than $2$ hours.

\begin{table*}
    \centering 
    \caption{Evaluation results of the ROMTrack~\citep{cai_robust_2023} attacked by the IoU approach~\citep{jia_iou_2021} on the UAV123~\citep{mueller_benchmark_2016} dataset and protocol for three different upper bounds on the added noise in normal direction up to $10$ processing steps.}
    \label{tab:uav}
    \begin{tabular}{p{0.05\textwidth}p{0.065\textwidth}p{0.065\textwidth}p{0.065\textwidth}p{0.065\textwidth}p{0.065\textwidth}p{0.065\textwidth}p{0.065\textwidth}p{0.065\textwidth}p{0.065\textwidth}p{0.065\textwidth}} \\ 
    \toprule
     & \multicolumn{3}{c}{AUC} & \multicolumn{3}{c}{Precision}  & \multicolumn{3}{c}{Norm Precision}  \\
    \midrule
    $\zeta$ & Original & Attack & Drop & Original & Attack & Drop &  Original & Attack & Drop  \\
    \midrule 
    %\multirow{3}{*}{ROMTrack}
     % & 5k & 69.74 & 68.09 & 2.36$\%$ & 90.83 & 89.26 & 1.73$\%$ & 85.30 & 84.01 & 1.51$\%$  \\
      8k & 69.74 & 66.85 & 4.14$\%$ & 90.83 & 89.31 & 1.67$\%$ & 85.30 & 83.00 & 2.70$\%$  \\
       10k & 69.74 & 65.46 & 6.14$\%$ & 90.83 & 87.81 & 3.32$\%$ & 85.30 & 81.73 & 4.18$\%$  \\
      12k & 69.74 &  63.61& 8.79$\%$ & 90.83 & 86.31 & 4.98$\%$ & 85.30 & 79.71 & 6.55$\%$  \\
       \bottomrule
    \end{tabular}
\end{table*}

The results of the attack on ROMTrack~\citep{cai_robust_2023} using the IoU method~\citep{jia_iou_2021} with different upper bounds are presented in Table~\ref{tab:uav}. It is clear that a higher upper bound leads to a more effective attack across all metrics. Despite the most substantial level of perturbation using the IoU method~\citep{jia_iou_2021} resulting in an $8.79\%$ decrease in the AUC metric, this outcome is insignificant. As shown in Figure~\ref{fig:iou}, increasing $\zeta$ generates a perceptible perturbation with a lower SSIM to the original frame, resulting in a noisier frame that damages the tracking performance of ROMTrack~\citep{cai_robust_2023} even more. However, the robust tracking performance is not affected more than $9\%$ per metric, even in the highest perturbation level. In other words, ROMTrack~\citep{cai_robust_2023} demonstrates good adversarial robustness against IoU attack on UAV123 dataset.

\subsection{Transformer versus Non-transformer Trackers}
\label{exp4}

In this experiment, we aim to study the adversarial robustness of trackers with different backbones especially transformer-based trackers compared to the non-transformer trackers.

\textbf{Evaluation Protocol}\quad This experiment is performed on the GOT10k dataset~\citep{huang_got-10k_2021} which uses OPE protocol and the results are reported for three metrics: Average Overlap (AO), Success Rate SR$_{0.5}$ with threshold $0.5$ and Success Rate SR$_{0.75}$ with threshold $0.75$. The GOT10k test set contains 180 video sequences with axis aligned bounding box annotations.

\textbf{Attacks Setting}\quad We applied a set of attacks including two white-box attacks, RTAA~\citep{jia_robust_2020}, SPARK~\citep{guo_spark_2020}, and two black-box attacks, IoU~\citep{jia_iou_2021} and CSA~\citep{yan_cooling-shrinking_2020}, against a set of trackers including three transformer trackers (i.e., TransT~\citep{chen_transformer_2021}, MixFormer~\citep{cui_mixformer_2022} and ROMTrack~\citep{cai_robust_2023}), two Siamese-based trackers (i.e., SiamRPN~\citep{li_high_2018} and DaSiamRPN~\citep{zhu_distractor-aware_2018}), and two discriminative trackers (i.e., DiMP~\citep{bhat_learning_2019} and PrDiMP~\citep{danelljan_probabilistic_2020}). As it is discussed in Attack Setups Section~\ref{transfer} and Table~\ref{tab:trans}, some attack methods are not applicable to some trackers due to the unavailability of the input for the attack algorithms. The RTAA and SPARK are used as a form of white-box attack, while the IoU and CSA methods are black-box attacks. For MixFormer and ROMTrack trackers, the related pre-trained networks specialized for GOT10k dataset and their parameters are loaded and examined. For DaSiamesRPN tracker, the checkpoint released to test the OTB100~\citep{wu_object_2015} dataset is used in this experiment.

\textbf{Results}\quad Table~\ref{tab:gtest} indicates the vulnerability of the object trackers with different backbones, transformers vs. non-transformers, against the adversarial attacks. The tracking performance before the attacks is indicated as the "No Attack" scores for each tracker, while the evaluation metrics after the attack are shown under the attack names. The "Drop($\%$)" columns in Table~\ref{tab:gtest} represents the drop percentage of scores from the original tracker response per metrics, and it is calculated as $\big ( 100 \times [\text{original} - \text{after attack}]/\text{original} \big )$. The drop scores are calculated to demonstrate the attack performances in comparison to each other and to assess the tracker's performance after each attack compared to other trackers.

\begin{table*}
    \centering 
    \caption{The performance of transformer~\citep{chen_high-performance_2023, cui_mixformer_2022, cai_robust_2023} and non-transformer trackers~\citep{li_high_2018, zhu_distractor-aware_2018, bhat_learning_2019, danelljan_probabilistic_2020} after white-box, SPARK~\citep{guo_spark_2020} and RTAA~\citep{jia_robust_2020}, and black-box, IoU~\citep{jia_iou_2021} and CSA~\citep{yan_cooling-shrinking_2020}, attacks on the GOT10k~\citep{huang_got-10k_2021} dataset.} % \todo{Never put vertical lines in tables.}}
    \label{tab:gtest}
    \begin{tabular}{p{0.14\textwidth}p{0.14\textwidth}p{0.08\textwidth}p{0.08\textwidth}p{0.08\textwidth}p{0.08\textwidth}p{0.08\textwidth}p{0.08\textwidth}} \\ 
    \toprule
     Tracker & Attacker & \multicolumn{3}{c}{Scores} & \multicolumn{3}{c}{Drop(\%)}    \\
     & & AO & SR$_{0.5}$ & SR$_{0.75}$ &  AO & SR$_{0.5}$ & SR$_{0.75}$ \\
    \midrule 
      \multirow{3}{*}{ROMTrack} 
        & No Attack & 0.729 & 0.830 & 0.702 & - & - & - \\
        & CSA & 0.716 & 0.814 & 0.682 & 1.78 & 1.93 & 2.85 \\
        & \textbf{IoU}  & 0.597 & 0.678 & 0.536 & \textbf{18.11} & \textbf{18.31} & \textbf{23.65} \\
      \midrule
      \multirow{5}{*}{TransT}
      & No Attack  & 0.723 & 0.823 & 0.682 &  - & - &  - \\
      & CSA   & 0.679 & 0.768 & 0.628  & 6.08 & 6.68 & 7.92 \\
      & IoU   & 0.532 & 0.611 & 0.432 & 26.42 & 25.76 & 36.66 \\
      & SPARK & 0.137 & 0.085 & 0.032 & 81.05 & 89.67 & 95.31 \\
      & \textbf{RTAA} &  0.048 & 0.019 & 0.011 & \textbf{93.36} & \textbf{97.69} & \textbf{98.39}\\
      \midrule
      \multirow{3}{*}{MixFormer}
       & No Attack & 0.696 & 0.791 & 0.656 & - & - & - \\
       & CSA   &  0.638 & 0.727 & 0.572 & 8.33 & 8.10 & 12.80 \\
      & \textbf{IoU} & 0.625 & 0.713 & 0.543 & \textbf{10.20} & \textbf{9.86} & \textbf{17.22} \\
      \midrule
      \multirow{3}{*}{PrDiMP}
      & No Attack  & 0.645 & 0.751 & 0.540 & - & - & - \\
      & \textbf{IoU} &  0.585 & 0.696 & 0.421  & \textbf{9.30} & \textbf{7.32} & \textbf{22.04} \\
      \midrule
      \multirow{3}{*}{DiMP}
      & No Attack  & 0.602 & 0.717 & 0.463 & - & - & - \\
      & \textbf{IoU} & 0.549 & 0.653 & 0.372 & \textbf{8.80} & \textbf{8.93} & \textbf{19.65} \\
      \midrule
      \multirow{5}{*}{SiamRPN}
      & No Attack  & 0.406 & 0.499 & 0.101 & - & - & - \\
      & CSA   & 0.382 & 0.460 & 0.083 & 5.91 & 7.81 & 17.82 \\
      & IoU   & 0.364 & 0.442 & 0.107 & 10.34 & 11.42 & -5.94 \\
      & SPARK &  0.279 & 0.353 & 0.059 & 31.28 & 29.26 & 41.58 \\
      & \textbf{RTAA} & 0.032 & 0.013 & 0.001 & \textbf{92.12} & \textbf{97.39} & \textbf{99.00} \\
      \midrule
       \multirow{5}{*}{DaSiamRPN}
      & No Attack  & 0.389 & 0.465 & 0.090 & - & - & - \\
      & CSA   &  0.314 & 0.353 & 0.061  & 19.28 & 24.09 & 32.22 \\
      & IoU   & 0.320 & 0.376 & 0.081 & 17.74 & 19.14 & 10 \\
      & SPARK & 0.255 & 0.313 & 0.053 & 34.44 & 32.69 & 41.11 \\
      & \textbf{RTAA} &  0.037 & 0.015 & 0.001  & \textbf{90.49} & \textbf{96.77} & \textbf{98.88} \\
       \bottomrule
    \end{tabular}
\end{table*}

 Beginning with the transformer trackers, the "No Attack" scores exhibit a relatively higher magnitude compared to non-transformer trackers. Within transformer trackers, the IoU and CSA methods are the mutual attacks, with the IoU method demonstrating a more significant damage compared to the CSA approach. Specifically, among transformer trackers, TransT appears to be particularly susceptible to perturbations induced by the IoU attack. However, a greater decline in performance for MixFormer following the CSA attack is observed in compared to other transformer trackers after the same attack, i.e. CSA attack. 

All the test attack approaches could be applied to TransT, SiamRPN, and DaSiamRPN trackers due to the availability of the classification and regression labels in these trackers. Intriguingly, the RTAA method provides the most powerful attacking performance against SiamRPN and DaSiamRPN and TransT trackers. Although the drop percentages of scores after SPARK attack is very close for Siamese-based trackers, this attack performance on TransT tracker is considerable. Repeated on for Siamese-based trackers, the IoU attack is resulted in greater performance drop rather than the CSA attack. 

Another intriguing observation is that MixFormer shows the smallest drop percentage after the IoU attack among transformer trackers. For example, the SR$_{0.5}$ of MixFormer drops by $9.86\%$, whereas the percentage drop of the same metric is $18.31\%$ for ROMTrack and $25.76\%$ for TransT tracker. This indicates that although ROMTrack and TransT initially achieve better scores before the attacks, their performance is more greatly affected by IoU attack, resulting in larger drops in their evaluation metrics. Respectively, ROMTrack, TransT and MixFormer are ranked in terms of adversarial robustness after both IoU and CSA attack per AO and SR$_{0.5}$ metrics among all trackers. However, for SR$_{0.75}$ measurements, the ranking of adversarial robustness is MixFormer, ROMTrack and TransT after IoU attack.

% \begin{table*}
%     \centering
%     \caption{The percentage drop after applying the IoU attack~\citep{jia_iou_2021} against transformer~\citep{cai_robust_2023,chen_high-performance_2023,jia_robust_2020} and non-transformer trackers~\citep{danelljan_probabilistic_2020,bhat_learning_2019,li_high_2018,zhu_distractor-aware_2018} on GOT10k~\citep{huang_got-10k_2021} validation set.}
%     \label{tab:iou}
%     \begin{tabular}{p{0.12\textwidth}p{0.1\textwidth}p{0.1\textwidth}p{0.1\textwidth}|p{0.1\textwidth}p{0.1\textwidth}p{0.1\textwidth}} \\ 
%     \toprule
%      & \multicolumn{3}{c}{AUC} & \multicolumn{3}{c}{Precision}    \\
%     \midrule
%      & Original & IoU  & Drop(\%) & Original &  IoU & Drop(\%) \\
%     \midrule 
%       ROMTrack  & 83.40 & 75.96 & 8.92 & 78.53 & 68.13 & 13.24   \\
%        TransT-SEG & 82.31 & 73.90 & 10.22 & 76.56 & 66.41 & 13.26 \\
%       MixFormer & 82.17 & 76.29 & 7.15 & 76.23 & 67.53 & 11.37 \\
%        \midrule
%       PrDiMP & 74.81 & 69.29 & 7.38 & 61.92 &  56.48 & 8.78 \\
%       DiMP & 73.21 & 69.08 & 5.64 & 58.66  & 53.14 & 9.41 \\
%       SiamRPN  & 54.67 & 51.68 & 5.47 & 22.22 & 21.63 & 2.65  \\
%       DaSiamRPN & 49.18  & 44.10 & 10.33 & 18.64 & 18.43 & 1.13 \\
%        \bottomrule
%     \end{tabular}
% \end{table*}

Conversely, the only applicable attack (from our attack methods) against discriminative trackers, DiMP and PrDiMP, is the black-box IoU method. A comparison between these two trackers reveals that PrDiMP exhibits greater robustness than DiMP. However, upon examining the evaluation scores, it's evident that PrDiMP experiences a more pronounced decline compared to DiMP in terms of percentage drop of SR$_{0.75}$ metric. As the percentage drop related to SR$_{0.75}$ for PrDiMP is $22.04\%$, while it is $19.65\%$ for the same score of DiMP tracker. For the drop percentage of SR$_{0.5}$ score, the PrDiMP, though, preserve its priority over DiMP tracker after the IoU attack. 

The TransT, ROMTrack and DaSiamRPN are the top three in drop percentages of AO and SR$_{0.5}$ scores after the IoU attack. The computed percentage drops indicates that although the transformer trackers demonstrate the highest scores before the attack, their scores after IoU attack fall more significantly than the non-transformer trackers in general. In some cases such as DaSiamRPN tracker, the percentage drop is also a big number per AO and SR$_{0.5}$ metrics.

%Another interesting observation is that the MixFormer shows smallest drop percentage after the IoU attack where the drop percentage is computed as calculated as $100 . (\text{original} - \text{after attack})/\text{original}$. For instance, the AUC of MixFormer is dropped by $7.15\%$, while the percentage drop of the same score is $8.92\%$ for the ROMTrack and $10.21\%$ for TransT-SEG trackers. It shows that although the ROMTrack and TransT-SEG presents better scores before the attacks, their performance more greatly tend to drop after the adversarial attacks. 

%On the other hand, the only applicable attack for discriminative trackers, DiMP and PrDiMP, is the black-box IoU method. Comparing these two trackers demonstrates that PrDiMP is the more robust tracker than DiMP. However, one can observe the evaluation scores of PrDiMP is declined more the DiMP in terms of percentage drop, i.e. $100*(\text{original} - \text{after attack})/\text{original}$. 
%It shows that although the PrDiMP is more robust tracker rather than DiMP in their original performance, its tracking results are more affected the adversarial attack. 

% Since the IoU attack is applied against all trackers, it is interesting to the check the adversarial robustness of trackers by computing the drop percentage. 
% For a better comparison, we created another table in order to compare the trackers performance after the IoU attack in terms of the drop percentage, Table~\ref{tab:iou}. 

%%%NEW TEST SET RESULTS

\section{Discussion}

Our investigation began in Section~\ref{exp1} with an examination of the adversarial robustness of transformer trackers in producing object bounding boxes compared to object binary masks. Our findings indicated that the prediction of binary masks was more susceptible to adversarial perturbations than object bounding box predictions, particularly in terms of the accuracy metric. Additionally, we observed that white-box attacks, specifically SPARK and RTAA, exhibited greater efficacy compared to black-box attacks such as IoU and CSA, when targeting the TransT-SEG tracker. Notably, among the transformer-based trackers analyzed, MixFormerM, which employs deeper relation modeling than TransT-SEG, demonstrated superior adversarial robustness in terms of computed EAO, accuracy, and robustness on the VOT2022STS dataset against a single attack. Furthermore, we observed that MixFormerM is not susceptible to attacks like SPARK and RTAA with its own gradients due to the absence of necessary attack proxies, namely classification and regression labels.

In the subsequent experiment (c.f. Section~\ref{exp2}), we demonstrated that the magnitude of perturbation shifts could influence or maintain the overall tracking outcomes, depending upon the attack strategy. An elevated level of perturbation consistently resulted in a higher count of super-perturbed search regions with increased L1 norm values for perturbations in both RTAA and SPARK. However, many perturbed search regions under the SPARK attack demonstrated the same values for SSIM and L1 norm metrics, indicating a stable imperceptibility and sparsity even with heightened levels of perturbation. 
 
For the third experiment (c.f. Section~\ref{exp3}), we evaluated the IoU attack across various upper bounds and discovered that, particularly in the context of black-box attacks involving random noises, the initialization point plays a crucial role. The IoU method becomes exceedingly time-consuming due to improper initialization. Higher upper bounds resulted in larger L1 norm values and smaller SSIM scores, suggesting less sparse and more perceptible noise in the perturbed frames.

In the concluding experiment (c.f. Section~\ref{exp4}), we examined the impact of various backbones---transformer-based, discriminative-based and Siamese-based---on visual tracking before and after adversarial attacks. Despite transformer trackers (ROMTrack, TransT, and MixFormer) showcasing the top-3 performance, their evaluation scores more notably decreased after applying the IoU method.

Another interesting finding is that the most effective attack may vary depending on the test set and tracker. Indeed, for TransT-SEG, the strongest attack on the VOT2022 dataset was SPARK (Section~\ref{exp1}), whereas RTAA outperformed SPARK on the GOT10k set on misleading TransT tracker's output (Section~\ref{exp4}). For the SiamRPN tracker, the similar trend on the GOT10k set was observed: RTAA outperformed SPARK.

In our settings, the only applicable attacks against MixFormer and ROMTrack are black-box attacks, i.e. IoU and CSA. Among black-box methods, the IoU attack outperformed CSA for TransT, MixFormer, MixFormerM and ROMTrack trackers, Sections~\ref{exp1} and~\ref{exp4}. However, the effect of the IoU method against ROMTrack and MixFormer's are trivial, Section~\ref{exp4}. The ROMTrack and MixFormer bounding box predictions were harmed by the IoU method up to $18.11\,\%$ and $18.81\%$ on GOT10k and VOT2022 datasets for the average overlap metric, respectively. This indicates that these trackers were not being challenged enough with existing applicable attack methods.

In addition to the aforementioned summary, our study also revealed the following observations:
\begin{itemize}[nosep,left=0pt]
    \item The generated perturbations for attacks that are applicable to transformer-based trackers have more impact on the object masks accuracy rather than on the accuracy of the bounding boxes on VOT2022ST dataset (Section~\ref{exp1}). 
    %\item For a specific output, object bounding box or object mask, the transformer tracker with stronger cross-attention modeling presents a more robust performance against the same attack. [Revise!!]
    \item Although it was demonstrated that adding previous perturbations to the current frame for perturbed search regions generation have an impact on the attacker performance~\citep{guo_spark_2020}, these previous perturbations result in more stable performance against changes in perturbation levels. For instance, such stability has been observed on SPARK~\citep{guo_spark_2020} attack performance against TransT tracker on the UAV123 dataset~\citep{mueller_benchmark_2016} (Section~\ref{exp2}). 
    \item The SPARK algorithm generates temporally sparse perturbations, meaning that the added perturbation to the search region is small for many frames. It results in imperceptible noise for those frames per video sequence, even though the perturbation level shifts to a higher value (Section~\ref{exp2}). 
    \item Increasing the perturbation level on SPARK results in more super-perturbed regions, i.e. regions with perceptible noise (Section~\ref{exp2}).
    %\todo{that last part of the sentence... what ``grows''? the number of super-perturbed regions? Earlier it said ``more'' which already implies it grows, it's redundant.} 
    \item In IoU attack approach~\citep{jia_robust_2020} and RTAA~\citep{jia_iou_2021} attack, adding a higher perturbation level generates more perceptible noise for all frames, which damage more the overall tracking performance (Sections~\ref{exp2} and~\ref{exp3}).
    \item The ranking of attack performance is sensitive to the experiment settings, dataset and protocol. For instance, SPARK method outperforms RTAA attack on VOT2022 for TransT-SEG tracker in Section~\ref{exp1}, while RTAA scores are smaller than SPARK scores for TransT tracker on UAV123 dataset (Section~\ref{exp2}). Again we observed that RTAA outperforms SPARK approaches in attacking TransT, SiamRPN and DaSiamRPN trackers on GOT10k set (Section~\ref{exp4}). 
    \item The outcome of the IoU attack is sensitive to its initialization. The evaluation process may take a long time due to unsuitable initialization point of the attack (Section~\ref{exp3}).  
    %\item The IoU attack, as the single mutual attack applicable to all trackers, results in a greater drop percentage on object trackers with stronger cross-attention and object modeling, such as ROMTrack and MixFormer, compared to those with lighter and independent layers of cross-attention, such as TransT and SiamRPN, on GOT10k dataset, Section~\ref{exp4}.  
    \item Although transformer trackers, containing ROMTrack, MixFormer and TransT, exhibits more robust performance before and after the adversarial attacks, comparing the percentage drops from original scores reveals that these tracker's average overlap decreased from the original scores greater than almost all other trackers after the IoU attack (Section~\ref{exp4}). 
    % \item The IoU attack, as the single mutual attack applicable to all trackers, results in a greater drop percentage on object trackers with stronger cross-attention and object modeling, such as ROMTrack and MixFormer, compared to those with lighter and independent layers of cross-attention, such as TransT and SiamRPN, on GOT10k~\citep{huang_got-10k_2021} dataset (Section~\ref{exp4}).  
    \item Discriminative trackers also demonstrate a great adversarial robustness and ranked immediately after the transformer trackers on GOT10k dataset (Section~\ref{exp4}). 
\end{itemize}

\section{Conclusion}

We conducted a study on adversarial attack methods for object trackers with the aim of testing their impact on transformer trackers and comparing their influences on visual trackers with different backbones. This paper includes several experiments on  various tracking datasets, trackers, and attack settings. We evaluated three transformer trackers, ranging from light to deep relation modeling, and four non-transformer trackers with Siamese-based and discriminative backbones. Four attack methods, including two in white-box and two in black-box settings, were employed to assess the adversarial robustness of visual trackers. Our results showed that the accuracy of binary masks are more likely to harm by the adversarial attack in comparison to the accuracy of predicted bounding boxes. We also discovered that changes in the perturbation level do not necessarily affect the tracking performance over a tracking dataset. The sparsity and imperceptibility of the perturbations can be managed by advising a proper loss function. We also found that transformer trackers' performances after the adversarial attack drops from original performance greatly when compared to Siamese-based and discriminative trackers after the same attack. Our study indicates the need for further research on the adversarial robustness of transformer trackers since the existing attacks did not challenge these trackers significantly. The white-box attacks against Siamese-based trackers are not applicable to the transformer and discriminative trackers using their gradients due to the change of tracker backbone and architecture. One potential path for future work is the development of novel white-box attacks to target these kinds of trackers. The other direction can be on focusing more effective black-box attacks since they do not depend on the tracker backbones. The insightful findings in the adversarial robustness of classification networks with transformer backbones are also worthy to transfer on tracking networks with transformer backbones.

\subsubsection*{Acknowledgments}
% This research was supported by funding from NSERC-Canada and MÉIE-Québec (DEEL project).

This work is supported by the DEEL Project CRDPJ 537462-18 funded by the Natural Sciences and Engineering Research Council of Canada (NSERC) and the Consortium for Research and Innovation in Aerospace in Québec (CRIAQ), together with its industrial partners Thales Canada inc, Bell Textron Canada Limited, CAE inc and Bombardier inc. \footnote{\url{https://deel.quebec}}

% Use unnumbered third level headings for the acknowledgments. All
% acknowledgments, including those to funding agencies, go at the end of the paper.
% Only add this information once your submission is accepted and deanonymized. 

\bibliography{references.bib}
\bibliographystyle{tmlr}

% \appendix
% \section{Appendix}
% You may include other additional sections here.

\end{document}